\documentclass[10pt,twocolumn,letterpaper]{article}

\usepackage{cvpr}              

\usepackage{graphicx}
\usepackage{amsmath}
\usepackage{amssymb}
\usepackage{booktabs}
\usepackage{paralist}
\usepackage{booktabs}       
\usepackage{xcolor}         
\usepackage{multirow}
\usepackage[normalem]{ulem}
\usepackage{float}
\usepackage{makecell}
\usepackage{algorithm}
\usepackage{algpseudocode}

\usepackage[pagebackref,breaklinks,colorlinks]{hyperref}

\usepackage[capitalize]{cleveref}
\crefname{section}{Sec.}{Secs.}
\Crefname{section}{Section}{Sections}
\Crefname{table}{Table}{Tables}
\crefname{table}{Tab.}{Tabs.}

\begin{document}

\title{FFHQ-UV: Normalized Facial UV-Texture Dataset for 3D Face Reconstruction}

\author{
Haoran Bai$^{1}$\footnotemark[1] ~~~~~ Di Kang$^{2}$ ~~~~~ Haoxian Zhang$^{2}$ ~~~~~ Jinshan Pan$^{1}$\footnotemark[2] ~~~~~ Linchao Bao$^{2}$\\
$^{1}$Nanjing University of Science and Technology ~~~~~ $^{2}$Tencent AI Lab
}

\twocolumn[{
\maketitle
\vspace{-4mm}
\begin{figure}[H]
\hsize=\textwidth
\centering
\includegraphics[width=2.0\linewidth]{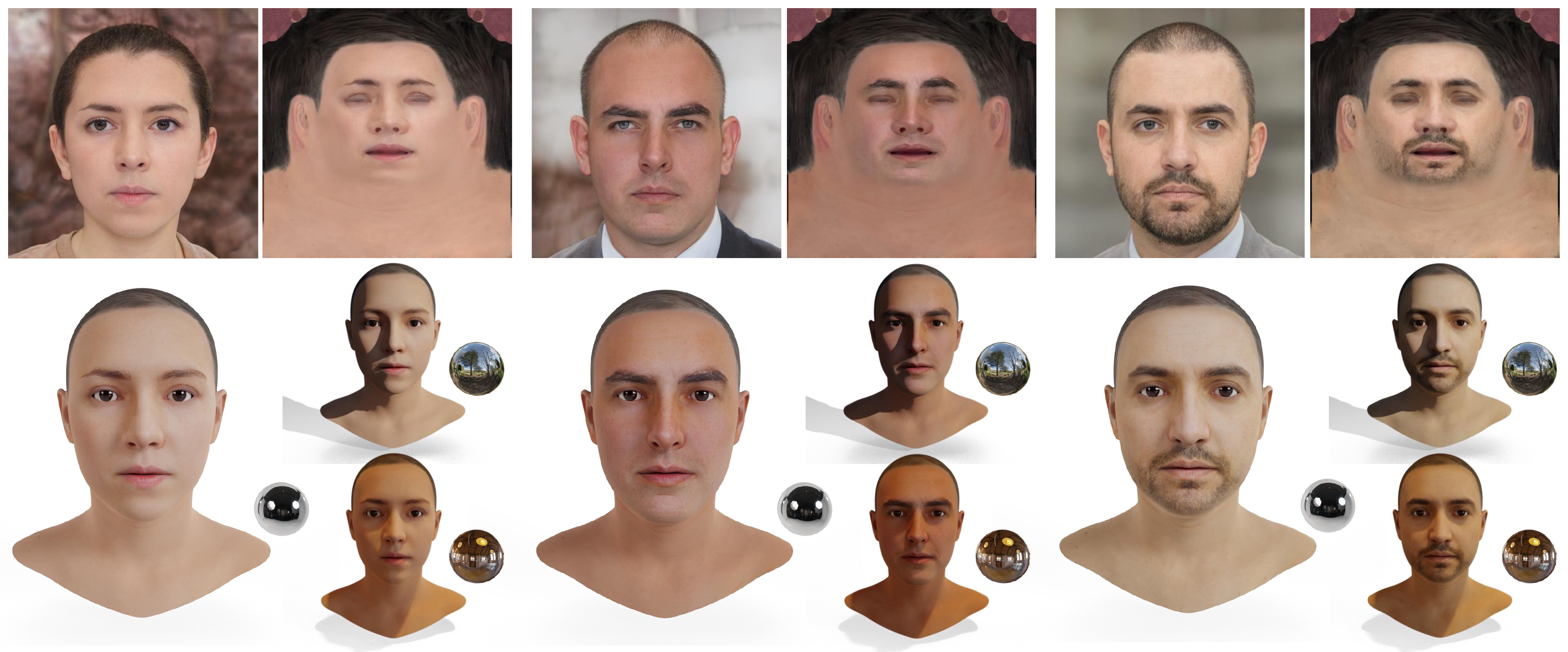}
\caption{%
\textbf{Examples of the proposed FFHQ-UV dataset.} 
From left-to-right and top-to-bottom are the normalized face images after editing, 
the produced texture UV-maps, and the rendered images under different lighting conditions. 
The proposed dataset is derived from FFHQ and preserves the most variations in FFHQ. 
The facial texture UV-maps in the proposed dataset are with even illuminations, neutral expressions, and cleaned facial regions (e.g. no eyeglasses and hair), which are ready for realistic renderings.
}
\label{fig:teaser}
\vspace{3mm}
\end{figure}
}]

\renewcommand{\thefootnote}{\fnsymbol{footnote}}
\footnotetext[1]{Work done during an internship at Tencent AI Lab.}
\footnotetext[2]{Corresponding author.}


\begin{abstract}
\vspace{-3mm}
We present a large-scale facial UV-texture dataset that contains over 50,000 high-quality texture UV-maps with even illuminations, neutral expressions, and cleaned facial regions, which are desired characteristics for rendering realistic 3D face models under different lighting conditions. 
The dataset is derived from a large-scale face image dataset namely FFHQ, with the help of our fully automatic and robust UV-texture production pipeline. 
Our pipeline utilizes the recent advances in StyleGAN-based facial image editing approaches to generate multi-view normalized face images from single-image inputs. 
An elaborated UV-texture extraction, correction, and completion procedure is then applied to produce high-quality UV-maps from the normalized face images. 
Compared with existing UV-texture datasets, our dataset has more diverse and  higher-quality texture maps. 
We further train a GAN-based texture decoder as the nonlinear texture basis for parametric fitting based 3D face reconstruction. 
Experiments show that our method improves the reconstruction accuracy over state-of-the-art approaches, and more importantly, produces high-quality texture maps that are ready for realistic renderings. 
The dataset, code, and pre-trained texture decoder are publicly available at \url{https://github.com/csbhr/FFHQ-UV}. 
\end{abstract}


\vspace{-4mm}
\section{Introduction}
\vspace{-1mm}
\label{sec:intro}

Reconstructing the 3D shape and texture of a face from single or multiple images is an important and challenging task in both computer vision and graphics communities.
Since the seminal work by Blanz and Vetter~\cite{blanz1999morphable} showed that the reconstruction can be effectively achieved by parametric fitting with a linear statistical model, namely 3D Morphable Model (3DMM), it has received active research efforts in the past two decades~\cite{egger20203d}.
While most 3DMM-based reconstruction approaches focused on improving the shape estimation accuracy, only a few works addressed the problem on texture UV-map recovery \cite{saito2017photorealistic,yamaguchi2018high,gecer2019ganfit,lattas2020avatarme,bao2021high,luo2021normalized,lee2020uncertainty}.

There are two key aspects that deserve attention in the texture map recovery problem, which are the \textit{fidelity} and the \textit{quality} of the acquired texture maps. 
In order to recover a high-fidelity texture map that better preserves the face identity of the input image, the texture basis in a 3DMM needs to have larger expressive capacities.
On the other hand, a higher-quality texture map requires the face region to be evenly illuminated and without undesired hairs or accessories, such that the texture map can be used as facial assets for rendering under different lighting conditions.

The method GANFIT \cite{gecer2019ganfit} trains a generative adversarial network (GAN) \cite{karras2017progressive} from 10,000 UV-maps as a texture decoder to replace the linear texture basis in 3DMM to increase the expressiveness.
However, their UV-maps in the training dataset are extracted from unevenly illuminated face images, and the resulting texture maps contain obvious shadows and are not suitable for differently lighted renderings. 
The same problem exists in another work \cite{lee2020uncertainty} based on UV-GAN \cite{deng2018uv}. 
The work AvatarMe \cite{lattas2020avatarme} combines a linear texture basis fitting with a super-resolution network trained from high-quality texture maps of 200 individuals under controlled conditions. 
HiFi3DFace \cite{bao2021high} improves the expressive capacity of linear texture basis by introducing a regional fitting approach and a detail refinement network, which is also trained from 200 texture maps. 
The Normalized Avatar work \cite{luo2021normalized} trains a texture decoder from a larger texture map dataset with over 5,000 subjects, consisting of high-quality scan data and synthetic data. 
Although the quality of the resulting texture maps of these methods is pretty high, the reconstruction fidelity is largely limited by the number of subjects in the training dataset. 
Besides, all these texture map datasets are not publicly available. 
A recent high-quality, publicly accessible texture map dataset is in the Facescape dataset \cite{yang2020facescape}, obtained in a controlled environment. However, the dataset only has 847 identities.

In this paper, we intend to contribute a large-scale, publicly available facial UV-texture dataset consisting of high-quality texture maps extracted from different subjects. 
To build such a large-scale dataset, we need a fully automatic and robust pipeline that can produce high-quality texture UV-maps from large-scale ``in-the-wild'' face image datasets. 
For the produced texture map, we expect it to have even illumination, neutral expression, and complete facial texture without occlusions such as hair or accessories.
This is not a trivial task, and there exist several challenges:
1) The uncontrolled conditions of the in-the-wild face images cannot provide high-quality normalized textures;
2) From a single-view face image, the complete facial texture cannot be extracted;
3) Imperfect alignment between the face image and the estimated 3D shape would cause unsatisfactory artifacts in the unwrapped texture UV-maps.

To address these issues, we first utilize StyleGAN-based image editing approaches~\cite{karras2020analyzing,abdal2021styleflow,shen2020interpreting} to generate multi-view normalized faces from a single in-the-wild image.
Then a UV-texture extraction, correction, and completion process is developed to fix unsatisfactory artifacts caused by imperfect 3D shape estimation during texture unwrapping, so that high-quality texture UV-maps can be produced stably.
With the proposed pipeline, we construct a large-scale normalized facial UV-texture dataset, namely FFHQ-UV, based on the FFHQ dataset~\cite{karras2019style}.
The FFHQ-UV dataset inherits the data diversity of FFHQ, and consists of high-quality texture UV-maps that can directly serve as facial assets for realistic digital human rendering (see Fig.~\ref{fig:teaser} for a few examples). 
We further train a GAN-based texture decoder using the proposed dataset, and demonstrate that both the fidelity and the quality of the reconstructed 3D faces with our texture decoder get largely improved.

In summary, our main contributions are:
\begin{compactitem}
    \item The first large-scale, publicly available normalized facial UV-texture dataset, namely FFHQ-UV, which contains over 50,000 high-quality, evenly illuminated facial texture UV-maps that can be directly used as facial assets for rendering realistic digital humans. 
    \item A fully automatic and robust pipeline for producing the proposed UV-texture dataset from a large-scale, in-the-wild face image dataset, which consists of StyleGAN-based facial image editing, elaborated UV-texture extraction, correction, and completion procedure.
    \item A 3D face reconstruction algorithm that outperforms state-of-the-art approaches in terms of both fidelity and quality, based on the GAN-based texture decoder trained with the proposed dataset.
\end{compactitem}

\section{Related Work}
\vspace{-1mm}

\noindent\textbf{3D Face Reconstruction with 3DMM.}
The 3D Face Morphable Model (3DMM) introduced by Blanz and Vetter~\cite{blanz1999morphable} represents a 3D face model with a linear combination of shape and texture basis, which is derived using Principal Component Analysis (PCA) from topologically aligned 3D face models. 
The task of reconstructing 3D face models from images is typically tackled by estimating the 3DMM parameters using either optimization-based fitting or learning-based regression approaches \cite{Zollhoefer2018FaceSTAR,egger20203d}. 
Beyond the PCA-based linear basis, various nonlinear bases emerged to enlarge the 3DMM representation capacity \cite{tran2018nonlinear,bouritsas2019neural,moschoglou20203dfacegan,gecer2019ganfit,jiang2019disentangled,wang2022faceverse,qiu2022sculptor}. 
With a neural network-based nonlinear 3DMM basis, the 3D face reconstruction turns into a task of finding the best latent codes of a mesh decoder \cite{bouritsas2019neural} or a texture decoder \cite{gecer2019ganfit}, which we still term as ``3DMM fitting''. 
For a thorough review of 3DMM and related reconstruction methods, please refer to the recent survey \cite{egger20203d}.

\vspace{1mm}
\noindent\textbf{Facial UV-Texture Recovery.}
While the texture basis in the original 3DMM~\cite{blanz1999morphable} is represented by vertex colors of a mesh, recent methods \cite{saito2017photorealistic,yamaguchi2018high,gecer2019ganfit,lattas2020avatarme,bao2021high,luo2021normalized,lee2020uncertainty, lin2022high, chung2022stylefaceuv} start to employ UV-map texture representation in order to fulfil high-resolution renderings. 
These methods can be categorized into two lines: image translation-based approaches \cite{saito2017photorealistic,yamaguchi2018high,lattas2020avatarme,bao2021high} or texture decoder-based approaches \cite{gecer2019ganfit,lee2020uncertainty,luo2021normalized}. 
The former approaches usually obtain a low-quality texture map and then perform an image translation to convert the low-quality texture map to a high-quality one \cite{saito2017photorealistic,yamaguchi2018high,lattas2020avatarme,bao2021high}. 
The latter approaches typically train a texture decoder as the nonlinear 3DMM texture basis and then employ a 3DMM fitting algorithm to find the best latent code for a reconstruction \cite{gecer2019ganfit,lee2020uncertainty,luo2021normalized}. 
Both the image translation operation and texture decoder need a high-quality UV-texture dataset for training. 
Unfortunately, to the best of our knowledge, there is no publicly available, high-quality facial UV-texture dataset in such a large scale that the data has enough diversity for practical applications (see Tab.~\ref{tab:uv-tex-dataset} for a summary of the sizes of the datasets used in literature).

\vspace{1mm}
\noindent\textbf{Facial Image Normalization.}
Normalizing a face image refers to the task of editing the face image such that the resulting face is evenly illuminated and in neutral expression and pose \cite{cole2017synthesizing,nagano2019deep}. 
In our goal of extracting high-quality, ``normalized'' texture maps from face images, we intend to obtain three face images in frontal/left/right views from a single image, such that the obtained images have even illumination, neutral expression, and no facial occlusions by forehead hairs or eyeglasses. 
To achieve this, we utilize recent advances in StyleGAN-based image editing approaches~\cite{abdal2021styleflow,shen2020interpreting,harkonen2020ganspace,patashnik2021styleclip}. 
In these approaches, an image is first projected to a latent code of an FFHQ-pretrained StyleGAN \cite{karras2019style,karras2020analyzing} model through GAN inversion methods \cite{richardson2021encoding,tov2021designing}, and then the editing can be performed in the latent space, where the editing directions can be discovered through the guidance of image attributes or labels.

\begin{figure*}[!t]
  \centering
   \includegraphics[width=0.98\linewidth]{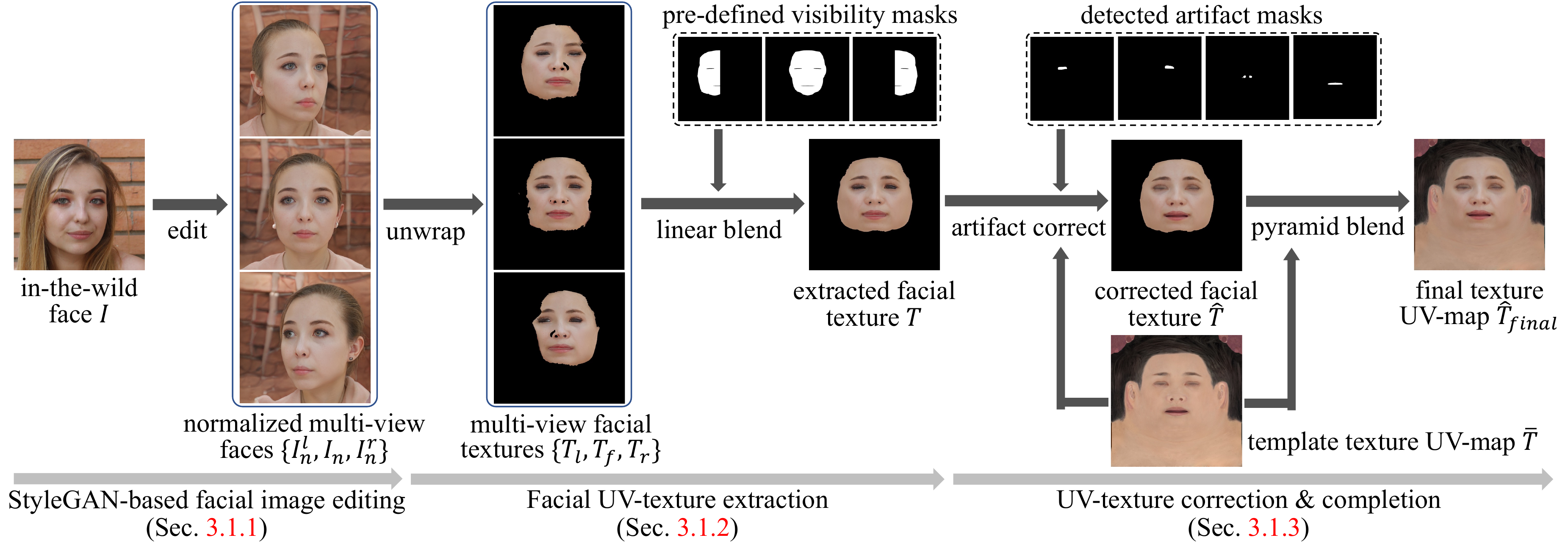}
   \caption{The proposed pipeline for producing normalized texture UV-map from a single in-the-wild face image, which mainly contains three modules: StyleGAN-based facial image editing, facial UV-texture extraction, and UV-texture correction \& completion.
   }
   \label{fig:dataset-pipeline}
\end{figure*}

\section{FFHQ-UV: Normalized UV-Texture Dataset}
\label{sec:dataset-pipeline}

In this section, we first describe the full pipeline for producing our normalized UV-texture dataset from in-the-wild face images (Sec. \ref{sec:datasetcreate}). 
Then we present extensive studies to analyze the diversity and quality of the dataset (Sec. \ref{sec:dataset-ana}). 

\begin{table}[!t]
\caption{Comparisons with existing UV-texture datasets, where $*$ denotes the dataset which is captured under controlled conditions.}
\small
\centering
\setlength{\tabcolsep}{0.5em}{
\begin{tabular}{l|cccc}
    \toprule
    \multirow{2}{*}{Datasets} & \multirow{2}{*}{\# images}   & \multirow{2}{*}{Resolution}  & Even & Public  \\
     &     &  & illum.  & avail. \\
    \midrule
    LSFM*~\cite{booth20163d}        & 10,000  & $512\times 512$  & $\times$ & $\times$ \\
    AvatarMe*~\cite{lattas2020avatarme} & 200  & $6144\times 4096$  & $\checkmark$ & $\times$ \\
    HiFi3DFace*~\cite{bao2021high}      & 200  & $2048\times 2048$  & $\checkmark$ & $\times$ \\
    Facescape*~\cite{yang2020facescape}      & 847   & $4096\times 4096$  & $\checkmark$ & $\checkmark$ \\
    \midrule
    WildUV~\cite{deng2018uv}        & 5,638  & $377\times 595$  & $\times$ & $\times$ \\
    NormAvatar~\cite{luo2021normalized}    & 5,601  & $256\times 256$  & $\checkmark$  & $\times$\\
    FFHQ-UV (Ours)                  & 54,165  & $1024\times 1024$ & $\checkmark$  & $\checkmark$\\
    \bottomrule
\end{tabular}
}
\label{tab:uv-tex-dataset}
\end{table}

\subsection{Dataset Creation}
\label{sec:datasetcreate}

The dataset creation pipeline, shown in Fig.~\ref{fig:dataset-pipeline}, consists of three steps: StyleGAN-based facial image editing (Sec.~\ref{sec:face-edit}), facial UV-texture extraction (Sec.~\ref{sec:tex-extract}), and UV-texture correction \& completion (Sec.~\ref{sec:tex-completion}).

\vspace{-2mm}
\subsubsection{StyleGAN-Based Facial Image Editing}
\label{sec:face-edit}
\vspace{-2mm}

To extract high-quality texture maps from in-the-wild face images, we first derive multi-view, normalized face images from single-view images, where the resulting face images have even illumination, neutral expression, and no occlusions (e.g., eyeglasses, hairs).
Specifically, we employ StyleFlow~\cite{abdal2021styleflow} and InterFaceGAN~\cite{shen2020interpreting} to automatically edit the image attributes in $\mathcal{W}+$ latent space of StyleGAN2~\cite{karras2020analyzing}. 
For each in-the-wild face image $I$, we first obtain its latent code $w$ in $\mathcal{W}+$ space using the GAN inversion method e4e~\cite{tov2021designing}, and then detect the attribute values of the inverted image $I_{inv}=G(w)$ from StyleGAN generator $G$, so that we can normalize these properties in the following semantic editing.
The attributes we intend to normalize include lighting, eyeglasses, hair, head pose, and facial expression. 
The lighting condition, represented as spherical harmonic (SH) coefficients, are predicted using DPR model~\cite{zhou2019deep}, and the other attributes are detected using Microsoft Face API~\cite{msapi}.

For the lighting normalization, we set the target lighting SH coefficients to keep only the first dimension and reset the rest dimensions to zeros, and then use StyleFlow~\cite{abdal2021styleflow} to get evenly illuminated face images. 
As in the SH representation, only the first dimension of the SH coefficients represents uniform lighting from all directions, while the other dimensions represent lighting from certain directions which are undesired. 
After the lighting normalization, we normalize the eyeglasses, head pose, and hair attributes by setting their target values to $0$, and obtain the edited latent code $w'$.
For the facial expression attribute, similar to InterFaceGAN~\cite{shen2020interpreting}, we find a direction $\beta$ of editing facial expression using SVM, and achieve the normalized latent code $\hat{w}$ by walking along the direction $\beta$ starting from $w'$.
To avoid over-editing, we further introduce an expression classifier to decide the stop condition for walking.
Here, we obtain the normalized face image $I_{n}=G(\hat{w})$.
Finally, two side-view face images $I_{n}^l$ and $I_{n}^r$ are generated using StyleFlow~\cite{abdal2021styleflow} by modifying the head pose attribute.

\vspace{-3mm}
\subsubsection{Facial UV-Texture Extraction}
\label{sec:tex-extract}
\vspace{-2mm}

The process of extracting UV-texture from a face image, also termed ``unwrapping'', requires a single-image 3D face shape estimator. 
We train a Deep3D model~\cite{deng2019accurate} with the recent 3DMM basis HiFi3D++~\cite{chai2022realy} to regress the 3D shape in 3DMM coefficients, as well as the head pose parameters, from each normalized face image. 
Then the facial UV-texture is unwrapped by projecting the input image onto the 3D face model. 
In addition, we employ a face parsing model~\cite{zllrunningfaceparsing} to predict the parsing mask for the facial region, so that non-facial regions can be excluded from the unwrapped texture UV-map.
In this way, for each face we obtain three texture maps, $T_f$, $T_l$, and $T_r$ from frontal, left, and right views, respectively. 
To fuse them together, we first perform a color matching between them to avoid color jumps. 
The color matching is computed from $T_l$ and $T_r$ to $T_f$ for each channel in YUV color space: 
\begin{equation}
    \setlength{\abovedisplayskip}{3pt}  
    \setlength{\belowdisplayskip}{3pt}  
    T_a'=\frac{T_a-\mu(T_a)}{\sigma(T_a) \times \omega} \times \sigma(T_b) + \mu(T_b),
    \label{eq:match-color}
\end{equation}
where $T_b$, $T_a$, and $T_a'$ are the target texture, source texture, and color-matched texture, respectively; $\mu$ and $\sigma$ denote the mean and standard deviation; and $\omega$ is a hyper-parameter (set to $1.5$ empirically) used to control the contrast of the output texture.
Finally, the three color matched texture maps are linearly blended together using pre-defined visibility masks (see Fig.~\ref{fig:dataset-pipeline}) to get a complete texture map $T$.

\vspace{-4mm}
\subsubsection{UV-Texture Correction \& Completion}
\label{sec:tex-completion}
\vspace{-2mm}

The obtained texture map $T$ usually contains artifacts near eyes, mouth, and nose regions, due to imperfect 3D shape estimation.
For example, eyeball and mouth interior textures\footnote{In our texture UV-map, only eyelids and lips are needed since the eyeballs and mouth interiors are independent mesh models similar to the mesh topology used in HiFi3D++ \cite{chai2022realy}.} that are undesired would appear in the texture map if the eyelids and lips between the estimated 3D shape and image are not well aligned. 
While performing local mesh deformation according to image contents \cite{zhou2019deep} could fix the artifacts in some cases, we find it would still fail for many images when processing large-scale dataset. 
To handle these issues, we simply extend and exclude error-prone regions and fill the missing regions with a template texture UV-map $\bar{T}$ (see Fig.~\ref{fig:dataset-pipeline}).
Specifically, we extract eyeball and mouth interior regions predicted from the face parsing result, and then unwrap them to the UV coordinate system after a dilation operation to obtain the masks of these artifacts $M_{leye}$, $M_{reye}$, and $M_{mouth}$ on its texture UV-map.
As for the nose region, we extract the nostril regions by brightness thresholding around the nose region to obtain a mask $M_{nostril}$, since nostril regions are usually dark. 
Then the regions in these masks are filled with textures from template $\bar{T}$ using Poisson editing~\cite{perez2003poisson} to get a texture map $\hat{T}$.

Finally, to obtain a complete texture map beyond facial regions (e.g., ear, neck, hair, etc.), we fill the rest regions using template $\bar{T}$, with a color matching using Eq.~\eqref{eq:match-color} followed by Laplacian pyramid blending~\cite{burt1983multiresolution}. 
The final obtained texture UV-map is denoted as $\hat{T}_{final}$.

\vspace{-3mm}
\subsubsection{FFHQ-UV Dataset}
\label{sec:ffhq-uv-dataset}
\vspace{-2mm}

We apply the above pipeline to all the images in FFHQ dataset~\cite{karras2019style}, which includes 70,000 high-quality face images with high variation in terms of age and ethnicity.
Images with facial occlusions that cannot be normalized are excluded using automatic filters based on face parsing results. 
The final obtained UV-maps are manually inspected and filtered, leaving 54,165 high-quality UV-maps at $1024 \times 1024$ resolution, which we name as FFHQ-UV dataset.
Tab.~\ref{tab:uv-tex-dataset} shows the statistics of the proposed dataset compared to other UV-map datasets.

\subsection{Ablation Studies}
\label{sec:dataset-ana}

We conduct ablation studies to demonstrate the effectiveness of the three major steps (i.e., StyleGAN-based image editing, UV-texture extraction, and UV-texture correction \& completion) of our dataset creation pipeline.
We compare our method to three baseline variants: 
1) ``w/o editing", which extracts facial textures directly from in-the-wild images without StyleGAN-based image editing; 
2) ``w/o multi-view", which uses only a single frontal-view face image for texture extraction; 
and 3) ``naive blending", which replaces our UV-texture correction \& completion step with a naive blending to the template UV-map.

\vspace{-4mm}
\subsubsection{Qualitative Evaluation}
\vspace{-1mm}

\begin{figure}[!t]
  \centering
   \includegraphics[width=0.98\linewidth]{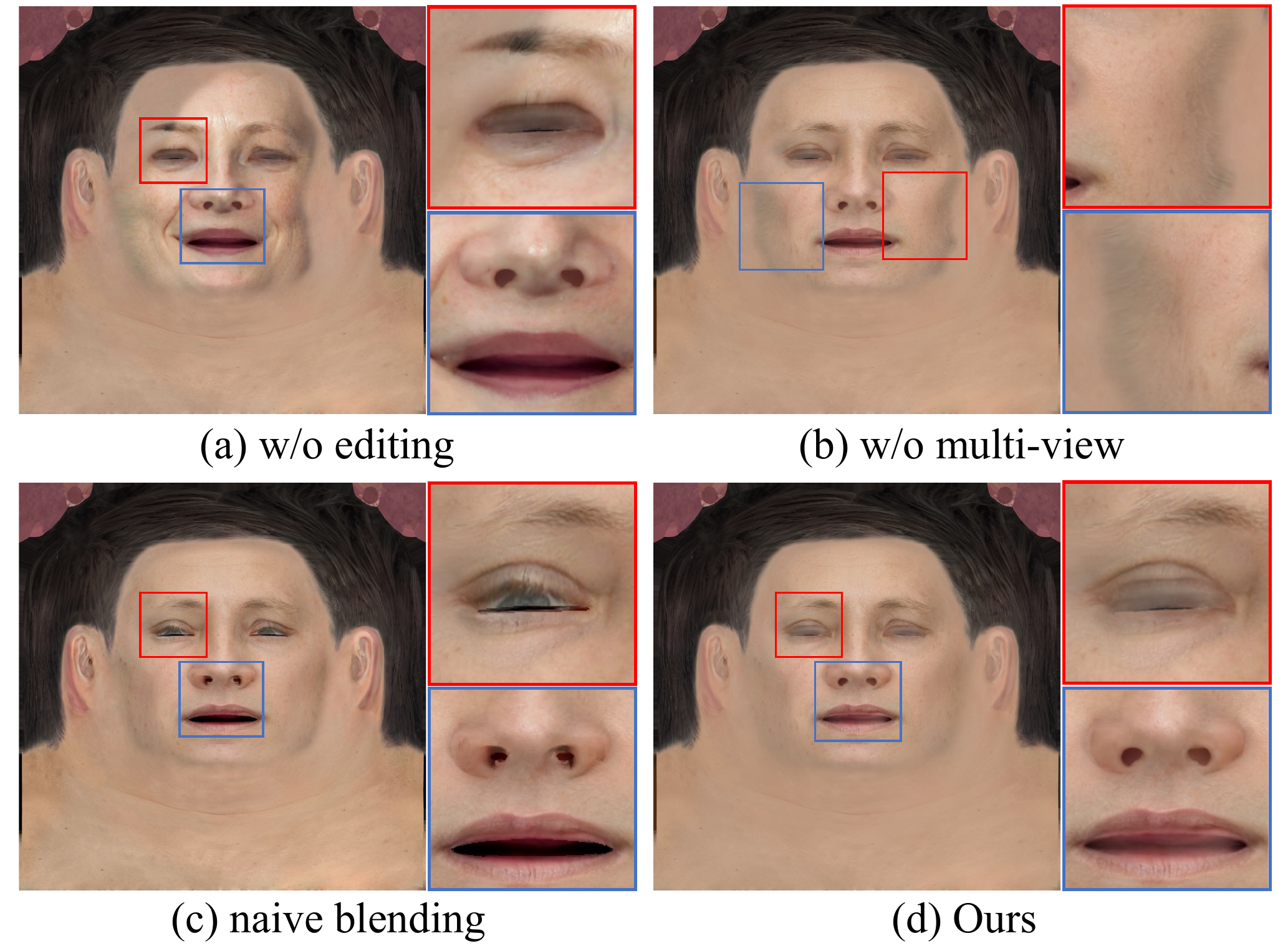}
   \caption{Extracted UV-maps with different variants of our pipeline. More results in the supplementary materials.}
   \label{fig:unwrap-ablation}
\end{figure}

Fig.~\ref{fig:unwrap-ablation} shows an example of the UV-maps obtained with different baseline methods. 
The baseline ``w/o editing"  (see Fig.~\ref{fig:unwrap-ablation}(a)) yields a UV-map that has significant shadows and occlusions of hair.
The UV-map generated by the baseline ``w/o multi-view" (see Fig.~\ref{fig:unwrap-ablation}(b)) contains smaller area of actual texture and heavily relies on template texture for completion.
For the baseline method ``naive blending", there are obvious artifacts near eyes, mouth, and nose regions (Fig.~\ref{fig:unwrap-ablation}(c)).
In contrast, our pipeline is able to produce high-quality UV-maps with complete facial textures (Fig.~\ref{fig:unwrap-ablation}(d)).

\vspace{-3mm}
\subsubsection{Data Diversity}
\vspace{-1mm}

\begin{table}[!t]
\caption{Quantitative evaluation on the diversity of the proposed dataset in terms of identity feature standard deviation, where all the values are divided by the value of FFHQ. $*$ indicates that ID features are extracted from rendered face images.}
\small
\centering
\setlength{\tabcolsep}{0.46em}{
\begin{tabular}{l|ccccc}
    \toprule
    Datasets & FFHQ  & \makecell{FFHQ \\-Inv}  & \makecell{FFHQ \\-Norm}  & \makecell{FFHQ \\-UV*}  & Facescape* \\
    \midrule
    ID std. & $100.0\%$  & $91.86\%$ & $90.06\%$ & $90.01\%$ & $84.24\%$ \\
    \bottomrule
\end{tabular}
}
\label{tab:id-std}
\end{table}

\begin{table}[!t]
\caption{Average identity similarity score between images in FFHQ-Norm and rendered images using FFHQ-UV. The score is averaged over the whole dataset. The ``negative samples'' is computed between the rendered image and an index-shifted real image. Note that the rendered images are in different views (with random head poses) from real images.}
\small
\centering
\setlength{\tabcolsep}{0.8em}{
\begin{tabular}{l|cccc}
    \toprule
    Methods & \makecell{negative \\samples}  & \makecell{w/o \\multi-view}  & \makecell{naive \\blending}  & Ours\\
    \midrule
    Similarity & 0.0648 & 0.7195 & 0.7712 & 0.7818 \\
    \bottomrule
\end{tabular}
}
\label{tab:fidelity-unwrap}
\end{table}

We expect FFHQ-UV would inherit the data diversity of FFHQ dataset~\cite{karras2019style}. 
To verify this, we compute the identity vector using Arcface~\cite{deng2019arcface} for each face, and then calculate the standard deviation of these identity vectors to measure the identity variations of the dataset. 
Tab.~\ref{tab:id-std} shows the identity standard deviation of the original dataset (FFHQ), the dataset inverted to the latent space (FFHQ-Inv), the normalized dataset using our StyleGAN-based facial image editing (FFHQ-Norm), the rendered face images using our facial UV-texture dataset (FFHQ-UV), where FFHQ-UV preserves the most identity variations in FFHQ (over $90\%$). 
Furthermore, our dataset has a higher identity standard deviation value compared to Facescape dataset~\cite{yang2020facescape}, indicating that FFHQ-UV is more diverse.
To analyze the identity preservation from FFHQ-Norm to FFHQ-UV, we compute the identity similarity between each image in FFHQ-Norm and the rendered image using the corresponding UV-map in FFHQ-UV with a random head pose. 
Tab.~\ref{tab:fidelity-unwrap} shows that the average identity similarity score over the whole dataset achieves $0.78$, which is pretty satisfactory considering that the rendered images are in different views (with random head poses) from real images. 
The table also shows that our result is superior to the results obtained by the baseline variants of our UV-map creation pipeline.

\vspace{-3mm}
\subsubsection{Quality of Even Illumination}
\vspace{-2mm}

\begin{figure}[!t]
  \centering
   \includegraphics[width=\linewidth]{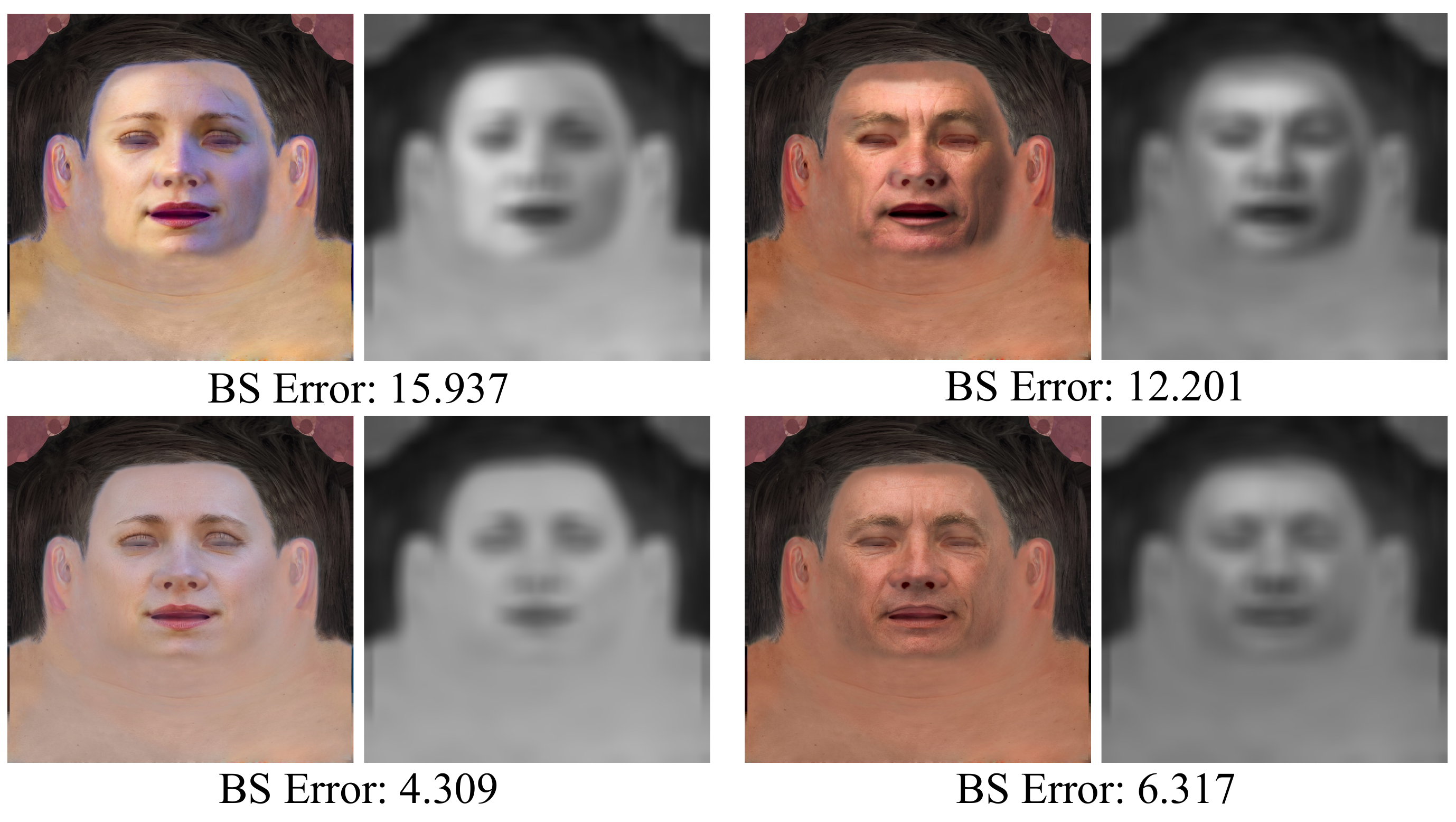}
   \caption{Examples of the computed BS Error on UV-maps and their intermediate results $\mathcal{B}_\alpha(T^Y)$. The first row are the texture UV-maps produced by the baseline method ``w/o editing", and the second row are those produced by the proposed method.}
   \label{fig:illumination-metric}
\end{figure}

\begin{table}[!t]
\caption{Quantitative evaluation on the illumination of the proposed UV-texture dataset in terms of BS Error, where $*$ denotes the dataset which is captured under controlled conditions.}
\small
\centering
\setlength{\tabcolsep}{1.1em}{
\begin{tabular}{l|ccc}
    \toprule
    Methods & Facescape*  & w/o editing  & Ours \\
    \midrule
    BS Error & 6.984  & 11.385  & 7.293 \\
    \bottomrule
\end{tabular}
}
\label{tab:illumination}
\end{table}

Having even illumination is one important aspect for measuring the quality of a UV-map~\cite{luo2021normalized}.
To quantitatively evaluate this aspect, we present a new metric, namely Brightness Symmetry Error (BS Error) as follows
\begin{equation}
    \setlength{\abovedisplayskip}{3pt}  
    \setlength{\belowdisplayskip}{3pt}  
    BS\_Error(T)=\left \| \mathcal{B}_\alpha(T^Y)-\mathcal{F}_h(\mathcal{B}_\alpha(T^Y))  \right \| _1,
    \label{eq:bs-error}
\end{equation}
where $T^Y$ denotes the $Y$ channel of $T$ in YUV space; $\mathcal{B}_\alpha(\cdot)$ denotes the Gaussian blurring operation with the kernel size of $\alpha$ (set to $55$ empirically); $\mathcal{F}_h(\cdot)$ denotes the horizontal flip operation.
The metric is based on the observation that an unevenly illuminated texture map usually has shadows on the face which makes the brightness of UV-map asymmetrical. 
Fig.~\ref{fig:illumination-metric} shows two examples of the computed BS Error on UV-maps.
Tab.~\ref{tab:illumination} shows the average BS Error computed over the whole dataset, which demonstrates that the StyleGAN-based editing step in our pipeline effectively improves the quality in terms of more even illumination.
In addition, the BS Error of our dataset is competitive with that of Facescape~\cite{yang2020facescape}, which is captured under controlled conditions with even illumination, indicating that our dataset is indeed evenly illuminated.

\begin{table}[!t]
\caption{Quantitative comparison of 3D face reconstruction on the RELAY benchmark~\cite{chai2022realy}. $*$ denotes the results reported from~\cite{chai2022realy}. ``FS" stands for Facescape dataset~\cite{yang2020facescape}.}
\label{tab:recons-realy}
\footnotesize
\resizebox{0.49\textwidth}{!}{
    \centering
    \begin{tabular}{l|cccc|c}
        \toprule
        Methods & nose  & mouth  & forehead  & cheek  & all\\
        \midrule
        3DDFA-v2*~\cite{guo2020towards} & 1.903  & 1.597 & 2.477  & 1.757  & 1.926 \\
        GANFit*~\cite{gecer2019ganfit} & 1.928  & 1.812 & 2.402  & 1.329  & 1.868 \\
        MGCNet*~\cite{shang2020self} & 1.771  & 1.417 & 2.268  & 1.639  & 1.774 \\
        Deep3D*~\cite{deng2019accurate} & 1.719  & 1.368 & 2.015  & 1.528  & 1.657 \\
        \midrule
        $\mathcal{N}_{enc}$ & \textbf{1.557}  & 1.661 & 1.940  & 1.014  & 1.543 \\
        PCA tex basis & 1.904  & 1.419 & 1.773  & 0.982  & 1.520 \\
        w/o multi-view & 1.780  & 1.419 & 1.711  & 0.980  & 1.473 \\
        \midrule
        w/ FS (scratch) & 1.731  & 1.653 & 1.711  & 1.207  & 1.576 \\
        w/ FS (finetune) & 1.570  & 1.576 & \textbf{1.581}  & 1.074  & 1.450 \\
        \midrule
        Ours & 1.681  & \textbf{1.339} & 1.631  & \textbf{0.943}  & \textbf{1.399} \\
        \bottomrule
    \end{tabular}
}
\end{table}

\begin{figure}[!t]
  \centering
   \includegraphics[width=\linewidth]{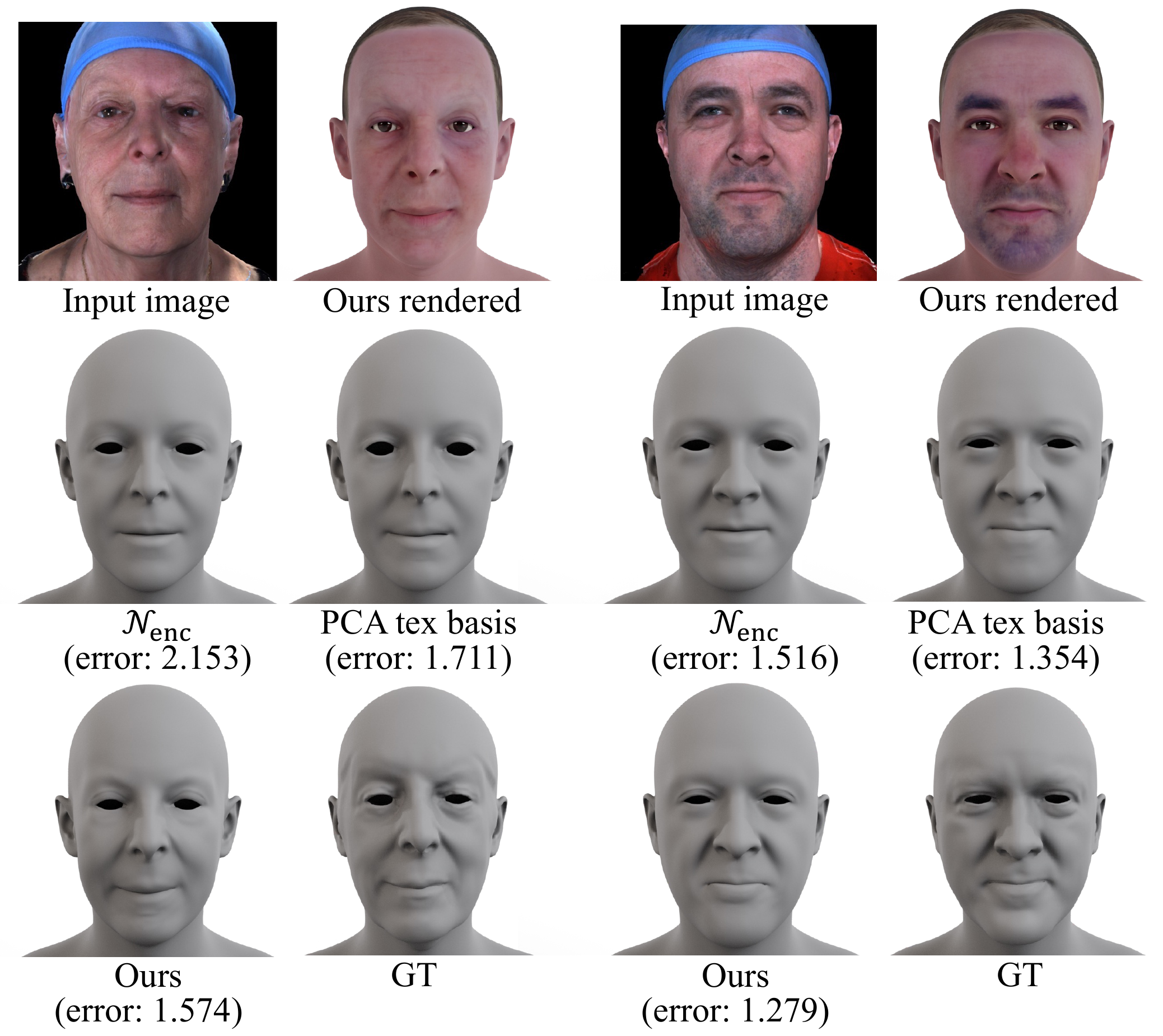}
   \caption{The shape reconstruction examples from REALY \cite{chai2022realy}, where our method reconstructs more accurate shapes and the rendered faces well resemble the input faces.}
   \label{fig:shape-acc}
\end{figure}

\begin{figure}[!t]
  \centering
   \includegraphics[width=\linewidth]{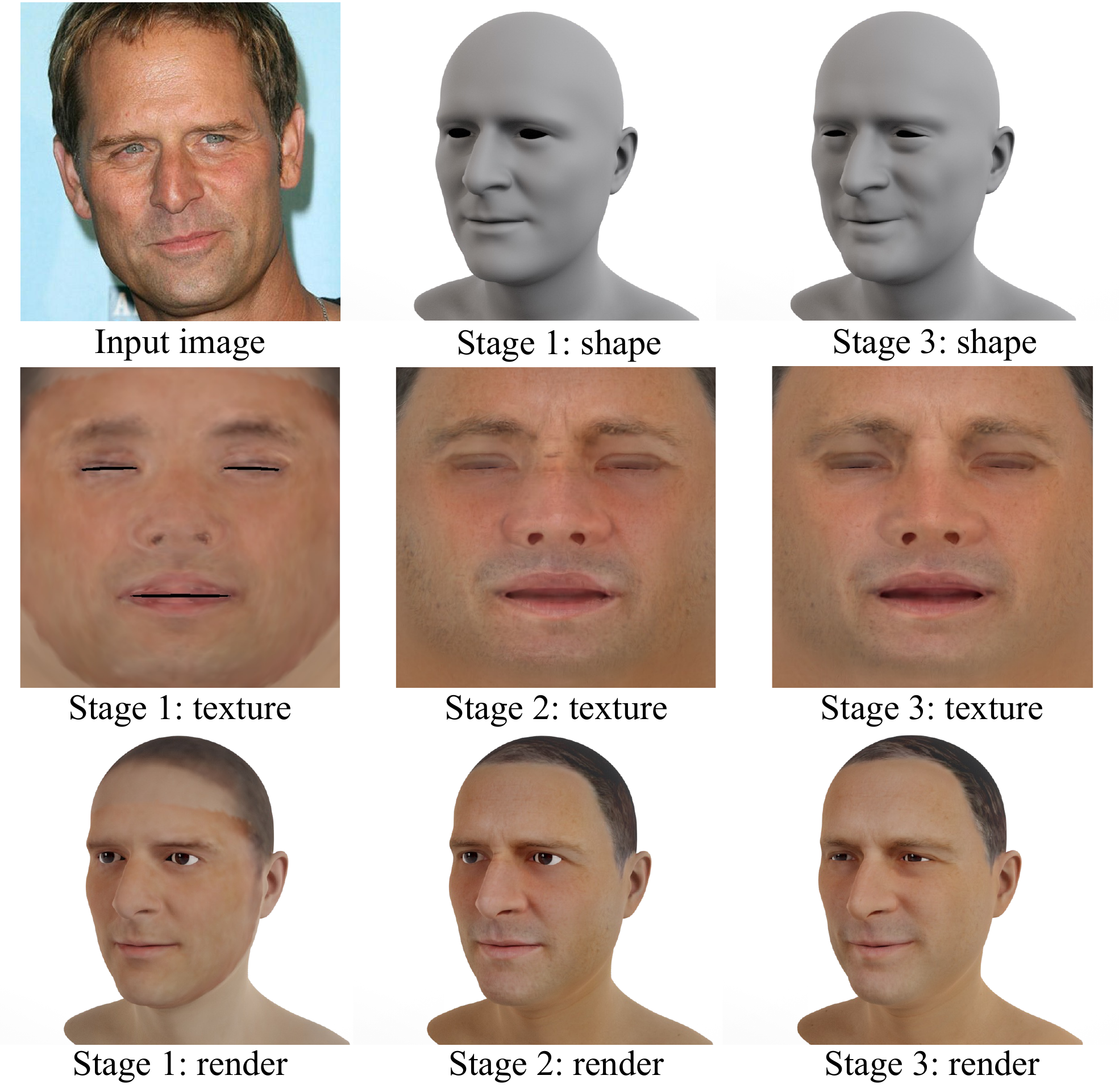}
   \caption{Intermediate reconstruction results of each stage in our algorithm. By comparing the rendered faces to the input face, it reveals that the final result better resembles the input face, especially the shape and texture around eyes, cheeks, and mouth regions.}
   \label{fig:fit-ablation}
\end{figure}

\begin{figure*}[!t]
  \centering
   \includegraphics[width=\linewidth]{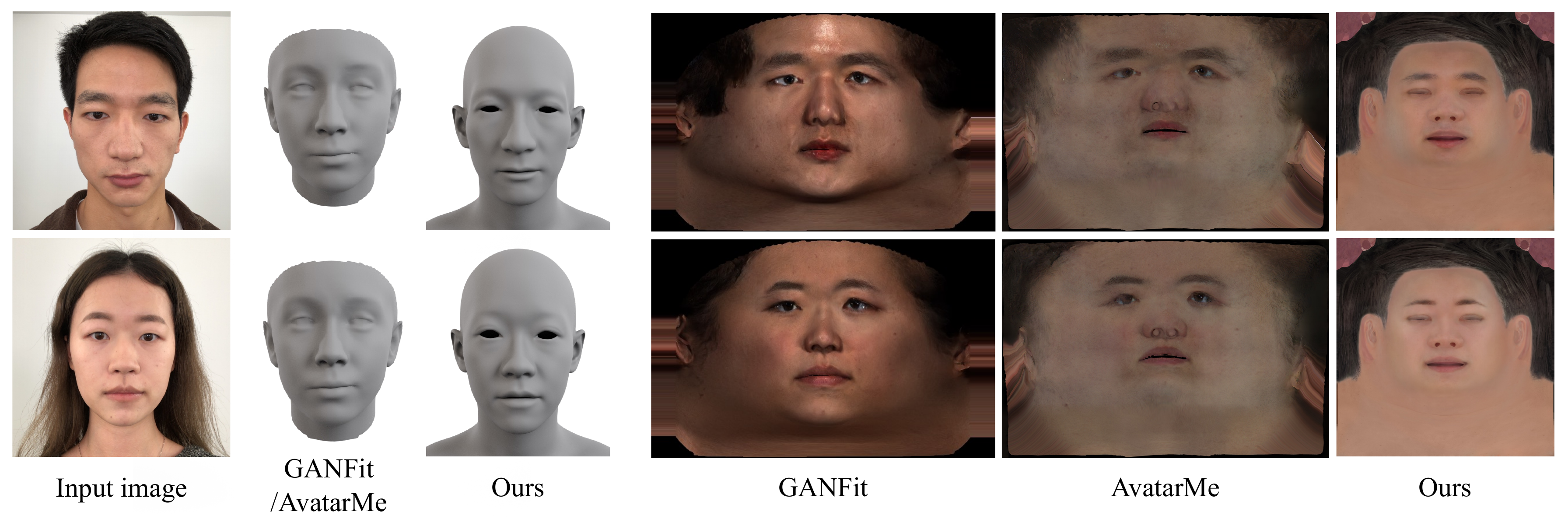}
   \caption{Visual comparison of the reconstruction results to state-of-the-art approaches GANFit~\cite{gecer2019ganfit} and AvatarMe~\cite{lattas2020avatarme}. Our reconstructed shapes are more faithful to input faces, and our recovered texture maps are more evenly illuminated and of higher quality.}
   \label{fig:sota-com-ganfit}
\end{figure*}

\vspace{2mm}
\section{3D Face Reconstruction with FFHQ-UV}
\vspace{-2mm}

In this section, we apply the proposed FFHQ-UV dataset to the task of reconstructing a 3D face from a single image, to demonstrate that FFHQ-UV improves the reconstruction accuracy and produces higher-quality UV-maps compared to state-of-the-art methods.

\vspace{-4mm}
\subsection{GAN-Based Texture Decoder}
\label{sec:tex-gan}
\vspace{-2mm}

We first train a GAN-based texture decoder on FFHQ-UV similar to GANFIT \cite{gecer2019ganfit}, using the network architecture of StyleGAN2 \cite{karras2020analyzing}. A texture UV-map $\tilde{T}$ is generated by
\begin{equation}
\setlength{\abovedisplayskip}{3pt}  
\setlength{\belowdisplayskip}{3pt}  
    \tilde{T}=\mathcal{G}_{tex}(z),
    \label{eq:tex-gan}
\end{equation}
where $z\in \mathcal{Z}$ denotes the latent code in the $\mathcal{Z}$ space and $\mathcal{G}_{tex}(\cdot)$ denotes the texture decoder.
The goal of our UV-texture recovery during 3D face reconstruction is to find a latent code $z^*$ that produces the best texture UV-map for an input image via the texture decoder $\mathcal{G}_{tex}(\cdot)$.

\subsection{Algorithm}
\label{sec:param-fit}

Our 3D face reconstruction algorithm consists of three stages: linear 3DMM initialization, texture latent code $z$ optimization, and joint parameter optimization. We use the recent PCA-based shape basis HiFi3D++ \cite{chai2022realy} and the differentiable renderer-based optimization framework provided by HiFi3DFace \cite{bao2021high}. The details are as follows.

\vspace{1mm}
\noindent\textbf{Stage 1: linear 3DMM initialization.}
We use Deep3D \cite{deng2019accurate} trained with shape basis HiFi3D++ \cite{chai2022realy} (the same as Sec.~\ref{sec:tex-extract}) to initialize the reconstruction. 
Given a single input face image $I^{in}$, the predicted parameters are $\left \{ p_{id}, p_{exp}, p_{tex}, p_{pose}, p_{light} \right \}$, where $p_{id}$ and $p_{exp}$ are the coefficients of identity and expression shape basis of HiFi3D++, respectively; $p_{tex}$ is the coefficient of the linear texture basis of HiFi3D++; $p_{pose}$ denotes the head pose parameters; $p_{light}$ denotes the SH lighting coefficients.
We use $\mathcal{N}_{enc}$ to denote the initialization predictor in this stage.

\vspace{1mm}
\noindent\textbf{Stage 2: texture latent code $z$ optimization.}
We use the parameters $\left \{ p_{id}, p_{exp}, p_{pose}, p_{light} \right \}$ initialized in the last stage and fix these parameters to find a latent code $z\in \mathcal{Z}$ of the texture decoder that minimizes the following loss:
\begin{equation}
    \setlength{\abovedisplayskip}{3pt}  
    \setlength{\belowdisplayskip}{3pt}  
    \mathcal{L}_{s2}= \lambda_{lpips}\mathcal{L}_{lpips}+\lambda_{pix}\mathcal{L}_{pix}+\lambda_{id}\mathcal{L}_{id}+\lambda_{reg}^z\mathcal{L}_{reg}^z,
    \label{eq:loss-s2}
\end{equation}
where $\mathcal{L}_{lpips}$ is the LPIPS distance~\cite{zhang2018unreasonable} between $I^{in}$ and the rendered face $I^{re}$; 
$\mathcal{L}_{pix}$ is the per-pixel $L_2$ photometric error between $I^{in}$ and $I^{re}$ calculated on the face region predicted by a face parsing model~\cite{zllrunningfaceparsing};
$\mathcal{L}_{id}$ is the identity loss based on the final layer feature vector of Arcface~\cite{deng2019arcface};
$\mathcal{L}_{reg}^z$ is the regularization term for the latent code $z$. 
Similar to~\cite{menon2020pulse}, we constrain the latent code $z$ on the hypersphere defined by $\mathcal{Z}'=\sqrt{d}S^{d-1}$ to encourage realistic texture maps, where $S^{d-1}$ is the unit sphere in $d$ dimensional Euclidean space.

\vspace{1mm}
\noindent\textbf{Stage 3: joint parameter optimization.}
In this stage, we relax the hyperspherical constraint on the latent code $z$ (to gain more expressive capacities) and jointly optimize all the parameters $\left \{ z, p_{id}, p_{exp}, p_{pose}, p_{light} \right \}$ by minimizing the following loss function:
\begin{equation}
\setlength{\abovedisplayskip}{3pt}  
\setlength{\belowdisplayskip}{3pt}  
    \begin{aligned}
    \mathcal{L}_{s3} = & \lambda_{pix}\mathcal{L}_{pix}
    + \lambda_{id}\mathcal{L}_{id} + \lambda_{reg}^z\mathcal{L}_{reg}^z  \\
    & + \lambda_{lm}\mathcal{L}_{lm} 
    + \lambda_{reg}^{id}\mathcal{L}_{reg}^{id}
    + \lambda_{reg}^{exp}\mathcal{L}_{reg}^{exp},
    \end{aligned}
    \label{eq:loss-s3}
\end{equation}
where $\mathcal{L}_{lm}$ denotes the 2D landmark loss with a 68-points landmark detector~\cite{bulat2017far};
$\mathcal{L}_{reg}^{id}$ and $\mathcal{L}_{reg}^{exp}$ are the regularization terms for the coefficients $p_{id}$ and $p_{exp}$.

\begin{figure}[!t]
  \centering
   \includegraphics[width=0.98\linewidth]{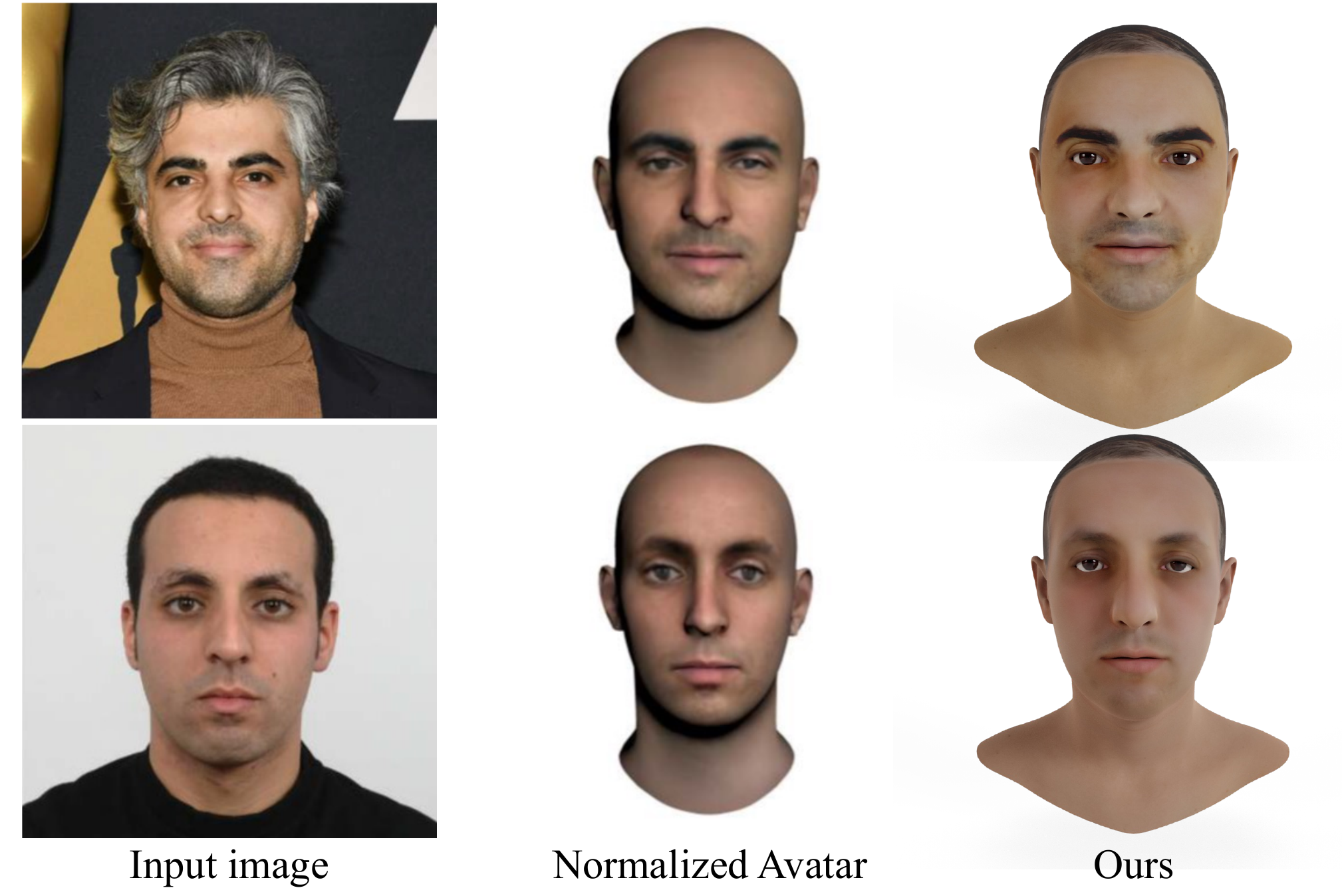}
   \caption{Visual comparison of reconstruction results to Normalized Avatar~\cite{luo2021normalized}. Our results better resemble the input faces.}
   \label{fig:sota-com-nas}
\end{figure}

\begin{figure*}[!t]
  \centering
   \includegraphics[width=0.98\linewidth]{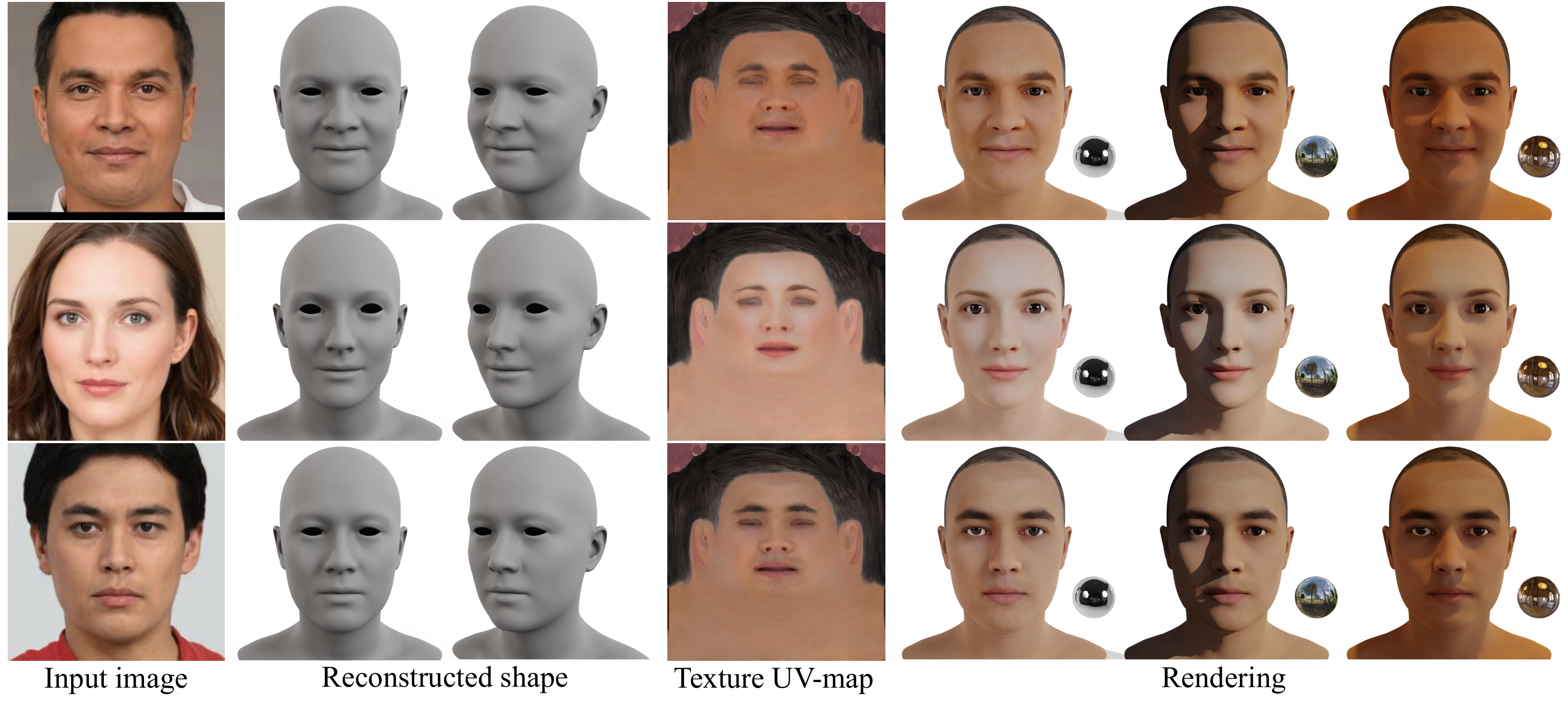}
   \caption{Examples of our reconstructed shapes, texture UV-maps, and renderings, where the produced textures are detailed and uniformly illuminated which can be rendered with different lighting conditions.}
   \label{fig:recons-relighting}
\end{figure*}

\vspace{-2mm}
\subsection{Evaluation}
\vspace{-1mm}

\noindent\textbf{Implementation details.}
The GAN-based texture decoder $\mathcal{G}_{tex}(z)$ is trained using the same hyperparameters as ``Config F" in StyleGAN2~\cite{karras2020analyzing} on 8 NVIDIA Tesla V100 GPUs, where the learning rate is set to $2e^{-3}$ and minibatch is set to $32$.
The loss weights $\{ \lambda_{lpips}, \lambda_{pix}, \lambda_{id}, \lambda_{reg}^z\}$ in Eq. \eqref{eq:loss-s2} of Stage 2 are set to $\{ 100, 10, 10, 0.05 \}$.
The loss weights $\{ \lambda_{pix}, \lambda_{id}, \lambda_{reg}^z, \lambda_{lm}, \lambda_{reg}^{id}, \lambda_{reg}^{exp}\}$ in Eq. \eqref{eq:loss-s3} of Stage 3 are set to $\{ 0.2, 1.6, 0.05, 2e^{-3}, 2e^{-4}, 1.6e^{-3}\}$.
We use Adam optimizer~\cite{kingma2014adam} to optimize 100 steps with a learning rate of 0.1 in Stage 2, and 200 steps with a reduced learning rate of 0.01 in Stage 3.
The total fitting time per image is around 60 seconds tested on an NVIDIA Tesla V100 GPU.

\vspace{1mm}
\noindent\textbf{Shape reconstruction accuracy.}
We first evaluate the shape reconstruction accuracy on REALY benchmark~\cite{chai2022realy}, which consists of 100 face scans and performs region-wise shape alignment to compute shape estimation errors.
Tab.~\ref{tab:recons-realy} shows the results, where our method outperforms state-of-the-art single-image reconstruction approaches including MGCNet~\cite{shang2020self}, Deep3D~\cite{deng2019accurate}, 3DDFA-v2~\cite{guo2020towards}, and GANFIT~\cite{gecer2019ganfit}. 
The table also shows the comparison to the results produced by the linear 3DMM initializer in Stage 1 ($\mathcal{N}_{enc}$), parameters optimization with linear texture basis instead of Stage 2 \& 3 (``PCA tex basis''), and the texture decoder trained using UV-map dataset created without generating multi-view images (``w/o multi-view''). 
The results demonstrate that our texture decoder effectively improves the reconstruction accuracy.
Fig. \ref{fig:shape-acc} shows two examples of the reconstructed meshes for visual comparison.
In addition, we adopt the texture UV-maps in Facescape~\cite{yang2020facescape}, which are carefully aligned to our topology with manual adjustment, to train a texture decoder using the same setting as ours.
Tab.~\ref{tab:recons-realy} (``w/ FS (scratch)") shows that training from scratch using Facescape does not perform well.
Using FFHQ-UV as pretraining and finetuning the decoder with Facescape brings substantial improvements (see ``w/ FS (finetune)"), but still decreases the result compared to ours, due to lost diversity.

\noindent\textbf{Texture quality.}
Fig. \ref{fig:fit-ablation} shows an example of the obtained UV-map in each stage and the corresponding rendered image. 
The result obtained from Stage 3 better resembles the input image, and the UV-map is more flatten and of higher quality. 
Fig. \ref{fig:sota-com-ganfit} shows two examples compared to GANFIT~\cite{gecer2019ganfit} and AvatarMe~\cite{lattas2020avatarme}, where our obtained meshes and UV-maps are superior to other results in terms of both fidelity and asset quality. 
Note that there are undesired shadows and uneven shadings in the UV-maps obtained by GANFIT and AvatarMe, while our UV-maps are more evenly illuminated.
Fig. \ref{fig:sota-com-nas} shows two examples of our results compared to Normalized Avatar~\cite{luo2021normalized}, where our rendered results better resemble the input faces, thanks to the more powerful expressive texture decoder trained on our much larger dataset. 
In Fig.~\ref{fig:recons-relighting}, we further show some examples of our reconstructed shapes, texture UV-maps, and renderings under different realistic lightings.
More results are presented in the supplementary materials.

\begin{table}[!t]
\caption{Fitting errors on CelebA-UV-100 with different texture decoders trained on different amounts of data from FFHQ-UV. The linear basis is the PCA-based texture basis in Stage 1 of Sec.~\ref{sec:param-fit}.}
\label{tab:inversion-lpips}
\small
\centering
\setlength{\tabcolsep}{0.5em}{
\begin{tabular}{l|cccc}
    \toprule
    Tex decoder & Linear basis  & 5,000  & 20,000  & 54,165 (Ours)\\
    \midrule
    LPIPS error & 0.4581 & 0.2029 & 0.1853 & \textbf{0.1487} \\
    \bottomrule
\end{tabular}
}
\end{table}

\noindent\textbf{Expressive power of texture decoder.}
We further validate the advantage of the expressive power of the texture decoder trained on larger datasets through the following experiments. 
We create a small validation UV-map dataset by randomly selecting 100 images from CelebA-HQ~\cite{karras2017progressive} and then creating their UV-maps using our pipeline in Sec.~\ref{sec:datasetcreate}. 
The validation dataset, namely CelebA-UV-100, consists of unseen data by texture decoders. 
We then use variants of texture decoders trained with different amounts of data to fit these UV-maps using GAN-inversion optimization \cite{karras2020analyzing}. 
Tab.~\ref{tab:inversion-lpips} shows the fitting results of the final average LPIPS errors between the target UV-maps and the fitted UV-maps.
The results show that the texture decoder trained on a larger dataset apparently has larger expressive capacities.

\vspace{-2mm}
\section{Conclusion and Future Work}
\vspace{-2mm}

We have introduced a new facial UV-texture dataset, namely FFHQ-UV, that contains over 50,000 high-quality facial texture UV-maps. 
The dataset is demonstrated to be of great diversity and high quality. 
The texture decoder trained on the dataset effectively improves the fidelity and quality of 3D face reconstruction. 
The dataset, code, and trained texture decoder will be made publicly available. 
We believe these open assets will largely advance the research in this direction, making 3D face reconstruction approaches more practical towards real-world applications.

\noindent\textbf{Limitations and future work.}
The proposed dataset FFHQ-UV is derived from FFHQ dataset \cite{karras2019style}, thus might inherit the data biases of FFHQ. 
While one may consider further extending the dataset with other face image datasets like CelebA-HQ \cite{karras2017progressive}, it may not help much because our dataset creation pipeline relies on StyleGAN-based face normalization, where the resulting images are projected to the space of a StyleGAN decoder, which is still trained from the FFHQ dataset. 
Besides, the evaluation of the texture recovery result lacks effective metrics that can reflect the quality of a texture map. 
It still requires visual inspections to judge which results are better in terms of whether the illuminations are even, whether facial details are preserved, whether there exist artifacts, whether the rendered results well resemble the input images, etc. 
We intend to further investigate these problems in the future.

\noindent\textbf{Acknowledgements.}
This work has been partly supported by the National Key R\&D Program of China (No. 2018AAA0102001), the National Natural Science Foundation of China (Nos. U22B2049, 61922043, 61872421, 62272230), and the Fundamental Research Funds for the Central Universities (No. 30920041109).

\clearpage
{\small
\bibliographystyle{ieee_fullname}
\bibliography{egbib}
}

\end{document}


\title{Supplemental Material for \\FFHQ-UV: Normalized Facial UV-Texture Dataset for 3D Face Reconstruction}

\author{Haoran Bai$^{1}$\footnotemark[1] ~~~~~ Di Kang$^{2}$ ~~~~~ Haoxian Zhang$^{2}$ ~~~~~ Jinshan Pan$^{1}$\footnotemark[2] ~~~~~ Linchao Bao$^{2}$\\
$^{1}$Nanjing University of Science and Technology ~~~~~ $^{2}$Tencent AI Lab
}

\maketitle

\renewcommand{\thefootnote}{\fnsymbol{footnote}}
\footnotetext[1]{Work done during an internship at Tencent AI Lab.}
\footnotetext[2]{Corresponding author.}


In this supplementary material, we first provide more analysis of the proposed data creation pipeline (Sec.~\ref{supp:sec:ana-pipeline}).
Then, we show more analysis of the proposed 3D face reconstruction algorithm, including quantitative evaluations, visual comparisons with state-of-the-art methods, and renderings under realistic conditions (Sec.~\ref{supp:sec:recons-cases}).
Finally, we show more examples of the proposed FFHQ-UV dataset rendered under different realistic lighting conditions (Sec.~\ref{supp:sec:unwrap-cases}).

\section{More analysis of the data creation}
\label{supp:sec:ana-pipeline}

As stated in Sec. 3.2 of the manuscript, we have analyzed the effectiveness of the three major steps (i.e., StyleGAN-based image editing, UV-texture extraction, and UV-texture correction \& completion) of our dataset creation pipeline.
In this supplemental material, we further provide more detailed analysis.

\subsection{Analysis on facial image editing}

\begin{figure}[h]
\centering
\includegraphics[width=\linewidth]{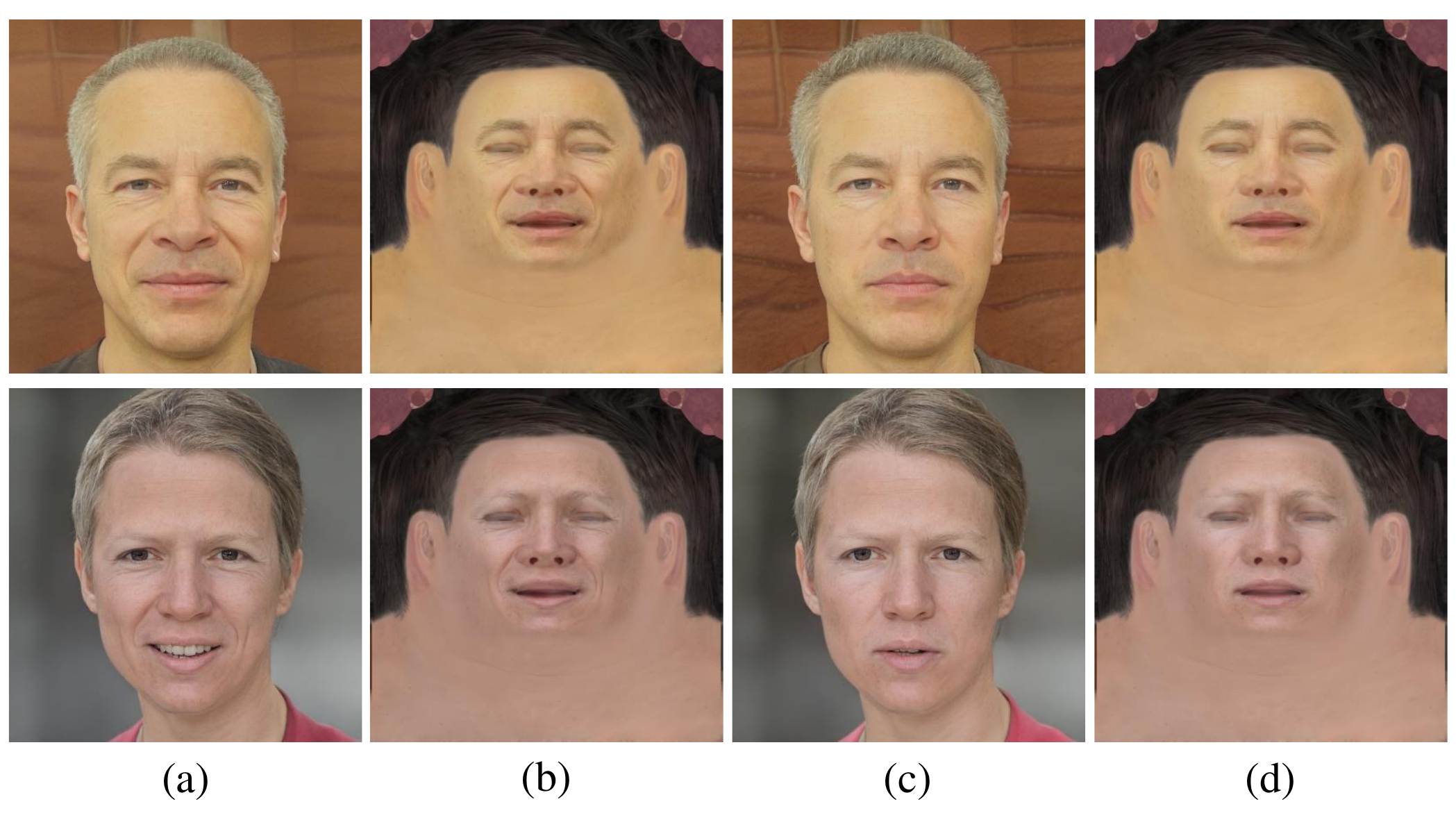}
\caption{Comparisons on the results of editing expression attribute using StyleFlow~\cite{abdal2021styleflow} and our method. 
Note that using StyleFlow cannot normalize the expression attribute well (a), resulting in texture UV-maps containing unwanted expression information (b).
In contrast, our method is able to generate neutral faces (c) and produce high-quality texture UV maps (d).}
\label{supp:fig:edit-exp}
\end{figure}

To obtain normalized faces, we first apply StyleFlow~\cite{abdal2021styleflow} to normalize the lighting, eyeglasses, hair, and head pose attributes of the input faces.
Then, we edit the facial expression attribute by walking the latent code along the found direction of editing facial expression.
One may wonder why not directly use StyleFlow to normalize the facial expression attribute.
To answer this question, we compare the method which uses StyleFlow to edit facial expression in Fig.~\ref{supp:fig:edit-exp}.
The results show that using StyleFlow cannot normalize the expression attribute well, resulting in texture UV-maps containing unwanted expression information.
In contrast, our method is able to generate neutral faces and produce high-quality texture UV maps.

\begin{figure*}[!t]
  \centering
   \includegraphics[width=0.98\linewidth]{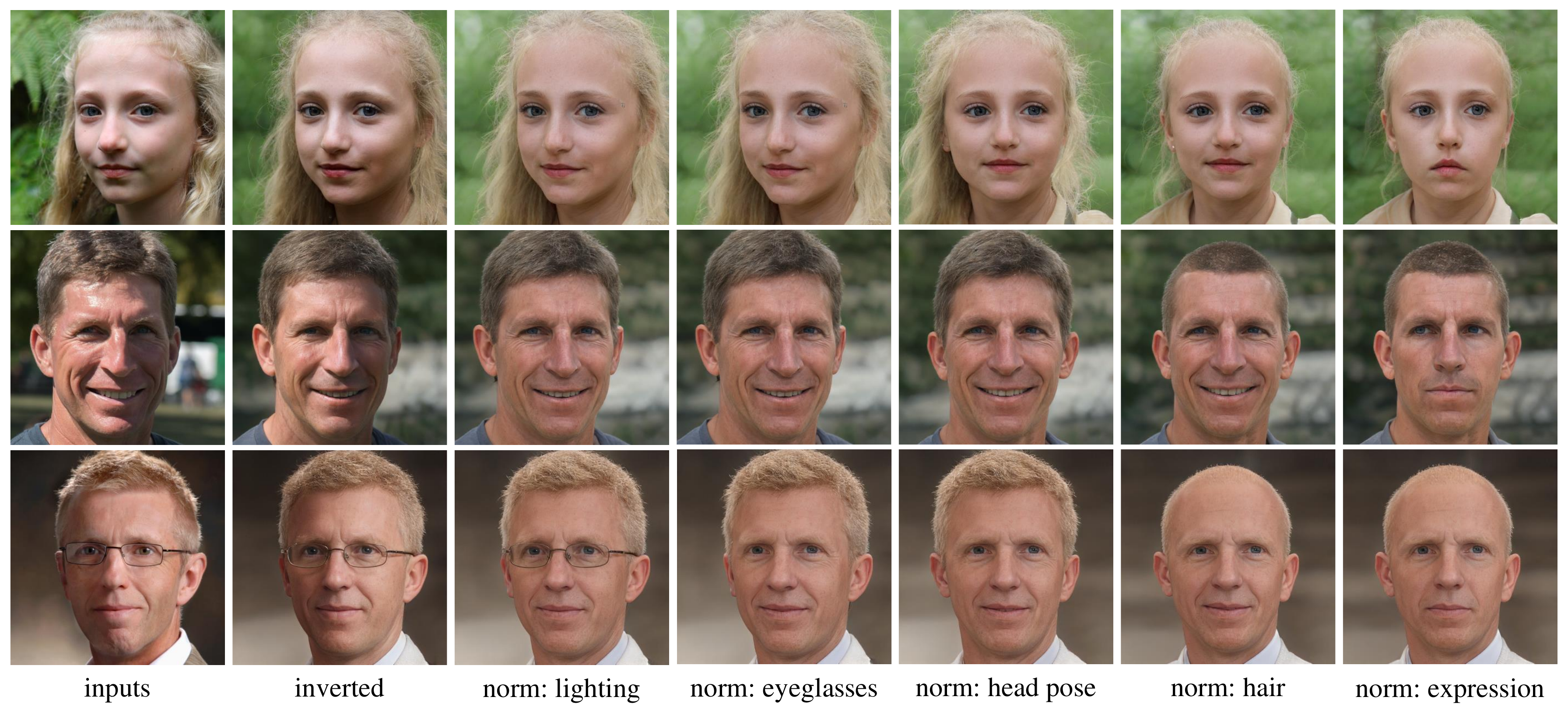}
   \caption{Intermediate results of StyleGAN-based facial image editing, where the lighting, eyeglasses, head pose, hair, and expression attributes of the input faces are normalized sequentially.}
   \label{supp:fig:editing-mid}
\end{figure*}

Furthermore, in Fig.~\ref{supp:fig:editing-mid}, we show some intermediate results of the proposed StyleGAN-based facial image editing.

\subsection{Analysis on UV-texture completion}

\begin{figure}[h]
  \centering
   \includegraphics[width=\linewidth]{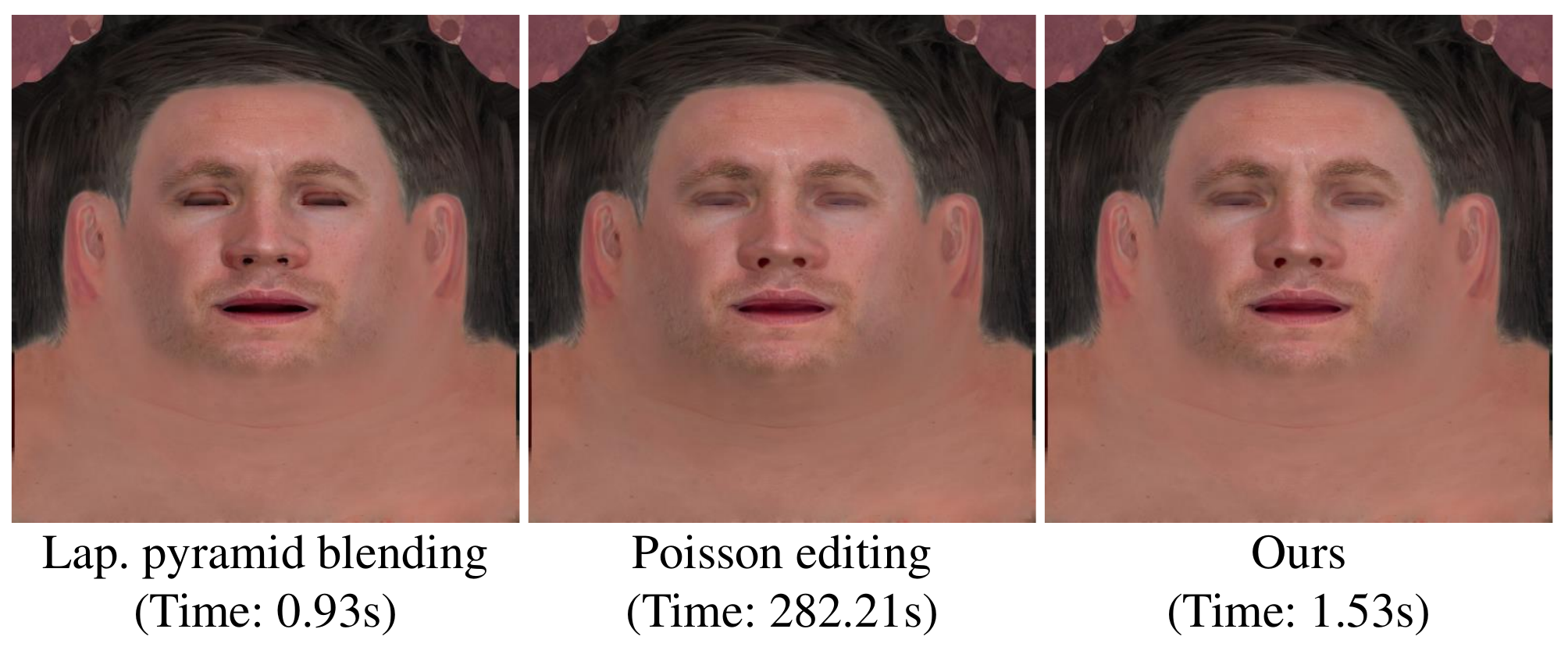}
   \caption{Comparisons on the results produced by only using Laplacian pyramid blending, only using Poisson editing, and using our method. Note that Laplacian pyramid blending cannot handle the facial regions (e.g., eyes, nostrils) well, and Poisson editing takes a huge amount of time to solve. In contrast, our method can achieve similar results to Poisson editing in a short time.}
   \label{supp:fig:blend-pyramid-poisson}
\end{figure}

After detecting the artifact masks, we first use Poisson editing~\cite{perez2003poisson} to correct the artifact regions and then use Laplacian pyramid blending~\cite{burt1983multiresolution} to handle the remaining non-face regions (e.g., ear, neck, hair, etc.).
One may wonder why not directly use Poisson editing or Laplacian pyramid blending to process all regions.
To answer this question, in Fig.~\ref{supp:fig:blend-pyramid-poisson}, we show the comparisons on the results produced by only using Laplacian pyramid blending, only using Poisson editing, and using the proposed method.
The results show that Laplacian pyramid blending cannot handle the facial regions (e.g., eyes, nostrils) well, because color matching operation usually fails in these regions.
Although Poisson editing can generate decent results, the huge time consumption is not suitable for creating large-scale datasets.
In contrast, the proposed method can achieve similar results to Poisson editing in a short time.

\subsection{More visual comparisons}

\begin{figure*}[!t]
  \centering
   \includegraphics[width=0.95\linewidth]{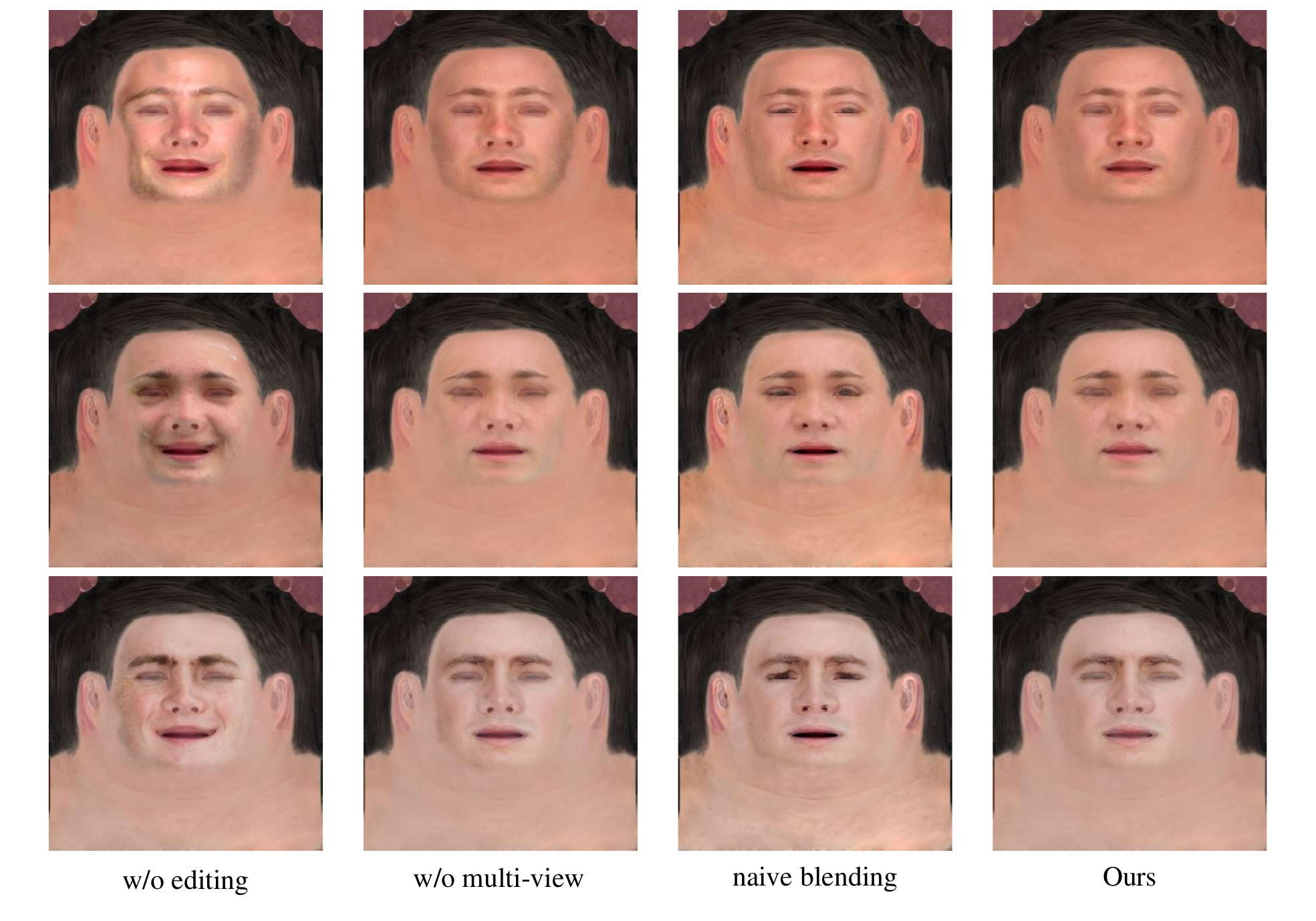}
   \caption{Extracted UV-maps by different baseline methods of our data creation pipeline, where our method is able to produce higher quality textures.}
   \label{supp:fig:unwrap-ablation}
\end{figure*}

In Sec. 3.2.1 of the manuscript, we have shown an example of the UV-maps obtained by different baseline methods.
In this supplemental material, we further provide more visual comparison in Fig.~\ref{supp:fig:unwrap-ablation}, which demonstrate the effectiveness of the three major steps of our dataset creation pipeline.

\subsection{Analysis on data diversity}

In Table 3 of the manuscript, we have shown the identity similarity computed between each image in FFHQ-Norm and the rendered image using the corresponding UV-map in FFHQ-UV with a random pose.
In this supplemental material, we further compute the average identity similarity with images in the original FFHQ dataset in Tab.~\ref{supp:tab:id-similarity} (``w/ orig. FFHQ"), showing that our dataset creation pipeline preserves identity well.
In addition, one may wonder whether using random pose renderings to compute identity features affects the computation of similarity scores.
Thereby, we further provide the average identity similarity with frontal renderings in Tab.~\ref{supp:tab:id-similarity} (``w/ frontal pose"), where ours is still the best.

\begin{table}[!h]
    \caption{Average ID similarity score.}
    \small
    \centering
    \setlength{\tabcolsep}{0.60em}{
    \begin{tabular}{l|cccc}
        \toprule
        Methods & \makecell{negative \\samples}  & \makecell{w/o \\multi-view}  & \makecell{naive \\blending}  & Ours\\
        \midrule
        w/ orig. FFHQ & 0.0710 & 0.3723 & 0.4092 & \textbf{0.4262} \\
        \midrule
        w/ frontal pose & 0.0594 & 0.8366 & 0.8481 & \textbf{0.8520} \\
        \bottomrule
    \end{tabular}
    }
    \label{supp:tab:id-similarity}
\end{table}

\section{More analysis of 3D face reconstruction}
\label{supp:sec:recons-cases}

\subsection{Evaluation on Facescape dataset}

As stated in Sec. 4.3 of the manuscript, we have evaluated the shape reconstruction accuracy on REALY benchmark~\cite{chai2022realy}.
In this supplemental material, we further evaluate the proposed 3D face reconstruction algorithm on the recent public Facescape dataset~\cite{yang2020facescape} with corresponding 3D scans and ground-truth texture UV-maps that can be used to compute quantitative metrics.
We apply the first 100 subjects from Facescape, and randomly select one picture from multi-view faces as input to evaluate the shape and texture of the monocular reconstruction.
For shape accuracy, we compute the average point-to-mesh distance between the reconstructed shapes and the ground-truth 3D scans.
For texture evaluation, we use the interactive nonrigid registration tool to align the ground-truth texture UV-map in Facescape to our topology, and use the mean L1 error as the metric.

\begin{table}[!h]
\caption{Quantitative comparison of reconstructed shapes and textures on the FaceScape dataset~\cite{yang2020facescape}.}
\small
\centering
\setlength{\tabcolsep}{1.0em}{
\begin{tabular}{l|cc}
    \toprule
    Methods & \makecell{shape (mm) $\downarrow$ \\(point-to-mesh dist.)}  & \makecell{texture $\downarrow$ \\(L1 error)} \\
    \midrule
    $\mathcal{N}_{enc}$ & 1.537  & 0.1719 \\
    PCA tex. basis & 1.524  & 0.1706 \\
    \midrule
    w/o multi-view & 1.502  & 0.1433 \\
    Ours & \textbf{1.495}  & \textbf{0.1425} \\
    \bottomrule
\end{tabular}
}
\label{supp:tab:recons-facescape}
\end{table}

Tab.~\ref{supp:tab:recons-facescape} shows the quantitative comparisons on different results
directly predicted by Deep3D in stage 1 (denoted as ``$\mathcal{N}_{enc}$''),
generated using linear PCA texture basis instead of the GAN-based texture decoder in Stage 2 \& 3 (``PCA tex. basis''),
generated using a texture decoder trained on the UV-map dataset created without generating multi-view images (``w/o multi-view''),
and generated using the texture decoder trained on our final FFHQ-UV dataset (``Ours'').

The results indicate that the proposed 3D face reconstruction algorithm, based on the GAN-based texture decoder trained with the proposed FFHQ-UV dataset, is able to improve the reconstruction accuracy in terms of both shape and texture.

\subsection{Evaluation on illumination and facial ID}

In Sec. 3.2 of the manuscript, we have evaluated the illumination and facial identity preservation of the data creation.
In this supplemental material, we further evaluate these on the proposed 3D face reconstruction algorithm.
We test them on REALY benchmark~\cite{chai2022realy} in Tab.~\ref{supp:tab:more-eval-recons}.
We first show the illumination evaluation of the reconstructed UV-maps (see Tab.~\ref{supp:tab:more-eval-recons} ``BS Error"), then compute the identity similarity between the input faces and the rendered faces to verify whether the reconstructed face can preserve the identity (see Tab.~\ref{supp:tab:more-eval-recons} ``Similarity"). Our method outperforms the baseline variants.

\begin{table}[!h]
    \caption{More evaluations on 3D face reconstruction.}
    \label{supp:tab:more-eval-recons}
    \small
    \centering
    \setlength{\tabcolsep}{0.45em}{
    \begin{tabular}{l|ccccc}
        \toprule
        Methods & \makecell{$\mathcal{N}_{enc}$}  & \makecell{PCA\\tex basis}  & \makecell{w/o \\editing}    & \makecell{w/o \\multi-view}  & Ours\\
        \midrule
        BS Error & - & - & 8.850 & 4.683 & \textbf{4.322} \\
        \midrule
        Similarity & 0.7832 & 0.7946 & 0.7153 & 0.7980 & \textbf{0.8102} \\
        \midrule
        Exp. Acc. & 73\% & 74\% & 69\% & 78\% & \textbf{82\%} \\
        \bottomrule
    \end{tabular}
    }
\end{table}

\subsection{Evaluation on facial expression}

In this section, we further evaluate facial expression preservation using the expression classifier.
As the expressions of faces in REALY~\cite{chai2022realy} are all neutral, we further collect 100 faces with different expressions from the websites.
We calculate the consistency of the two predictions (i.e. the original input images and the rendered images using the reconstructed shape/texture) from the expression classifier.
Results in Tab.~\ref{supp:tab:more-eval-recons} (``Exp. Acc.") show that our reconstruction method can better preserve the expressions.

\subsection{More visual results}

\begin{figure*}[!t]
  \centering
   \includegraphics[width=\linewidth]{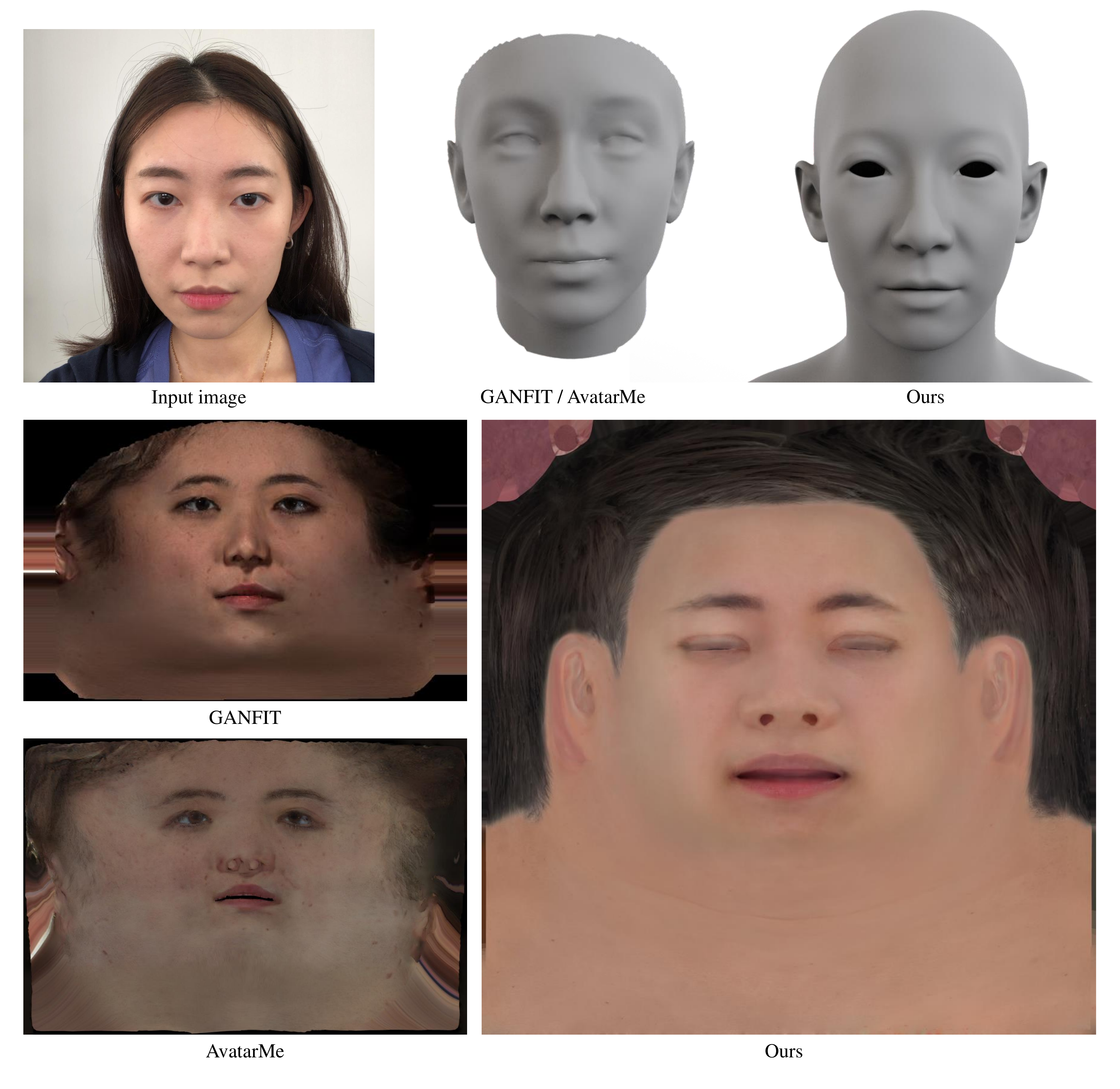}
   \caption{Visual comparisons of the reconstruction results to state-of-the-art approaches GANFIT~\cite{gecer2019ganfit} and AvatarMe~\cite{lattas2020avatarme}, where our reconstructed shape is more faithful to the input face.
   Note that there are undesired shadows and uneven shadings in the UV-maps obtained by GANFIT and AvatarMe, while our UV-map is more evenly illuminated and of higher quality.}
   \label{supp:fig:sota-ganfit-1}
\end{figure*}

\begin{figure*}[!t]
  \centering
   \includegraphics[width=\linewidth]{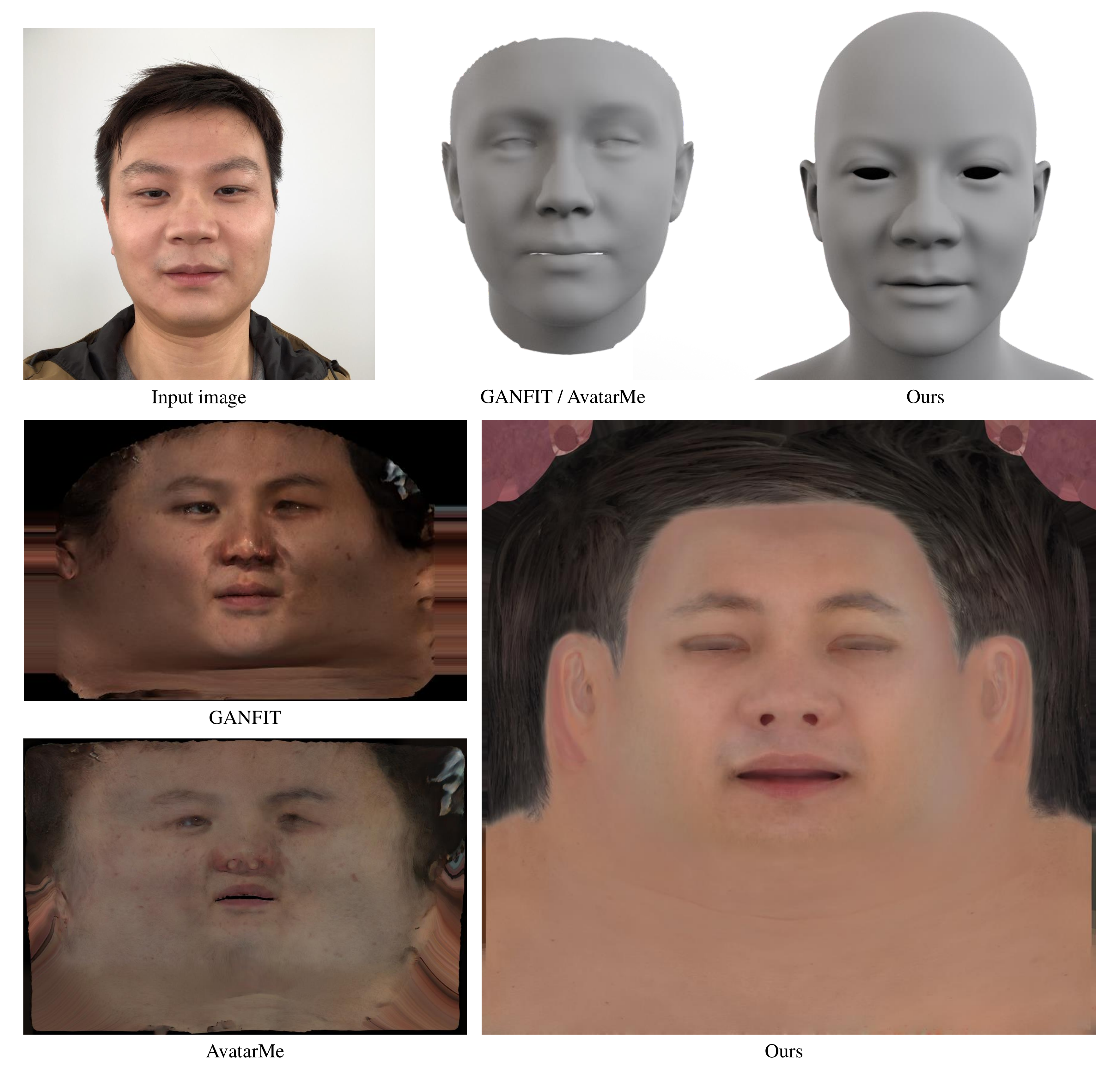}
   \caption{Visual comparisons of the reconstruction results to state-of-the-art approaches GANFIT~\cite{gecer2019ganfit} and AvatarMe~\cite{lattas2020avatarme}, where our reconstructed shape is more faithful to the input face (e.g., nose region).
   Note that there are undesired shadows and uneven shadings in the UV-maps obtained by GANFIT and AvatarMe, while our UV-map is more evenly illuminated and of higher quality.}
   \label{supp:fig:sota-ganfit-2}
\end{figure*}

\begin{figure*}[!t]
\centering
\includegraphics[width=0.9\linewidth]{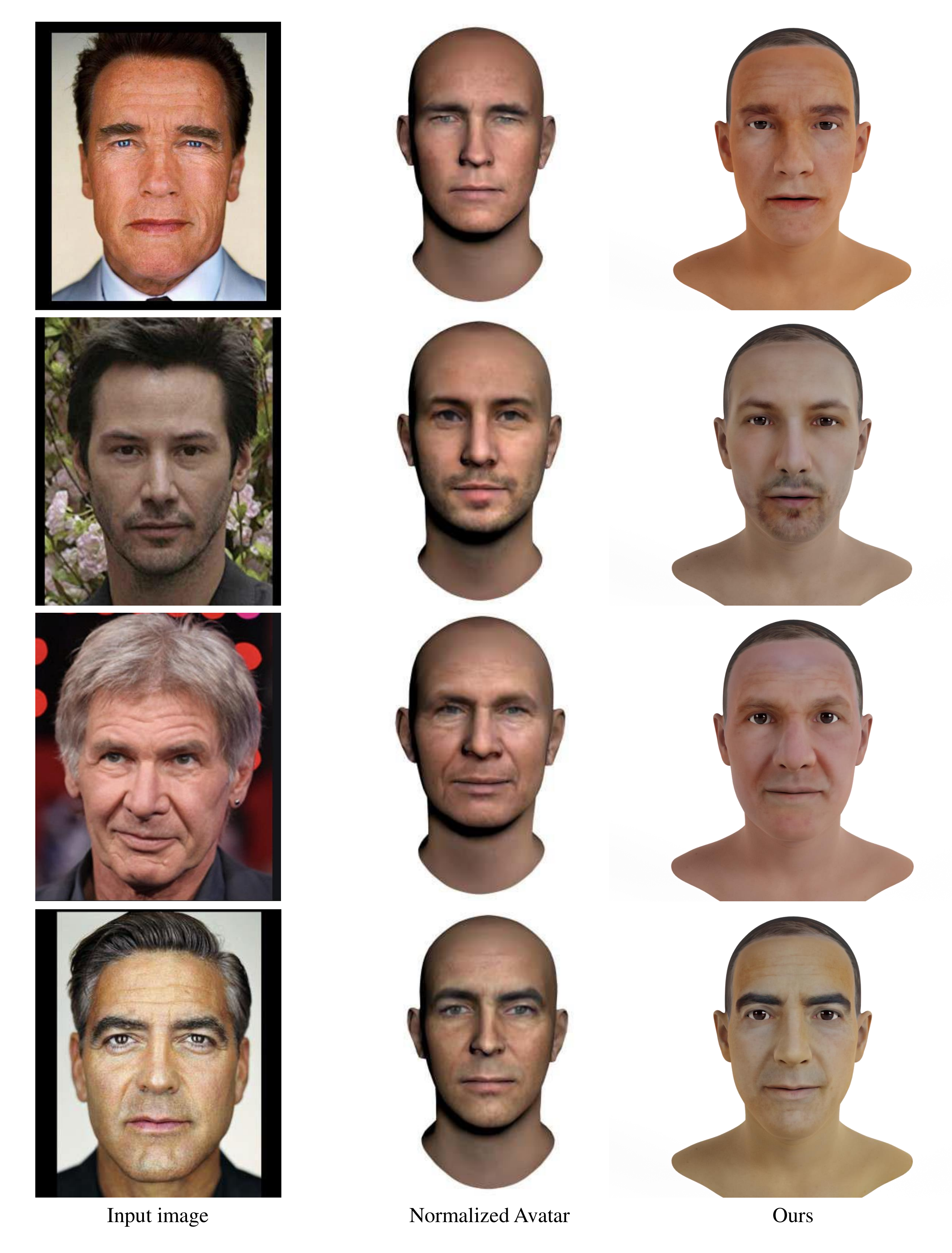}
\caption{Visual comparisons of the reconstructions between Normalized Avatar~\cite{luo2021normalized} and ours, where our results better resemble the input faces.
Note that our method is able to express more skin tones, thanks to the more powerful expressive texture decoder trained on our much larger dataset.}
\label{supp:fig:sota-nas}
\end{figure*}

\begin{figure*}[!t]
\centering
\includegraphics[width=0.9\linewidth]{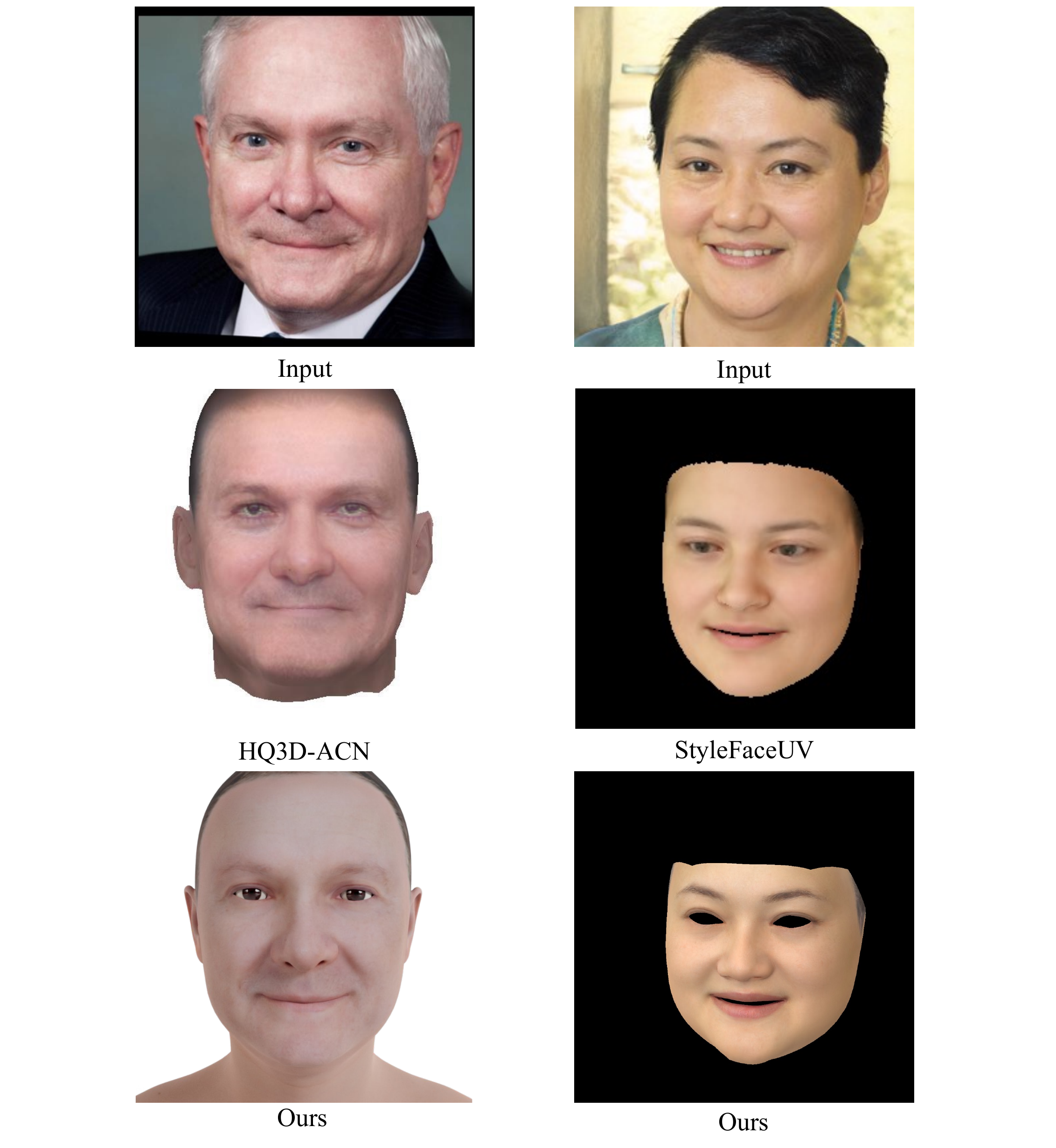}
\caption{Visual comparisons of the reconstruction results to recent approaches HQ3D-ACN~\cite{lin2022high} and StyleFaceUV~\cite{chung2022stylefaceuv}, where our results better resemble the input faces.}
\label{supp:fig:sota-hq3d-stylefaceuv}
\end{figure*}

In this section, we show more visual comparisons of the reconstructed faces produced by different methods, including GANFIT~\cite{gecer2019ganfit}, AvatarMe~\cite{lattas2020avatarme}, Normalized Avatar~\cite{lattas2020avatarme}, HQ3D-ACN~\cite{lin2022high}, StyleFaceUV~\cite{chung2022stylefaceuv}, and our method.
As shown in Fig.~\ref{supp:fig:sota-ganfit-1}-\ref{supp:fig:sota-ganfit-2}, meshes and texture UV-maps reconstructed by our method are superior to other results in terms of both fidelity and quality.
Compared to Normalized Avatar~\cite{lattas2020avatarme}, as shown in Fig.~\ref{supp:fig:sota-nas}, our results better resemble the input faces, and our method is able to express more skin tones, thanks to the more powerful expressive texture decoder trained on our much larger dataset.
In Fig.~\ref{supp:fig:sota-hq3d-stylefaceuv}, we show the visual comparisons of the reconstruction results to recent approaches HQ3D-ACN~\cite{lin2022high} and StyleFaceUV~\cite{chung2022stylefaceuv}, where our results better resemble the input faces. 

\begin{figure*}[!t]
  \centering
   \includegraphics[width=0.95\linewidth]{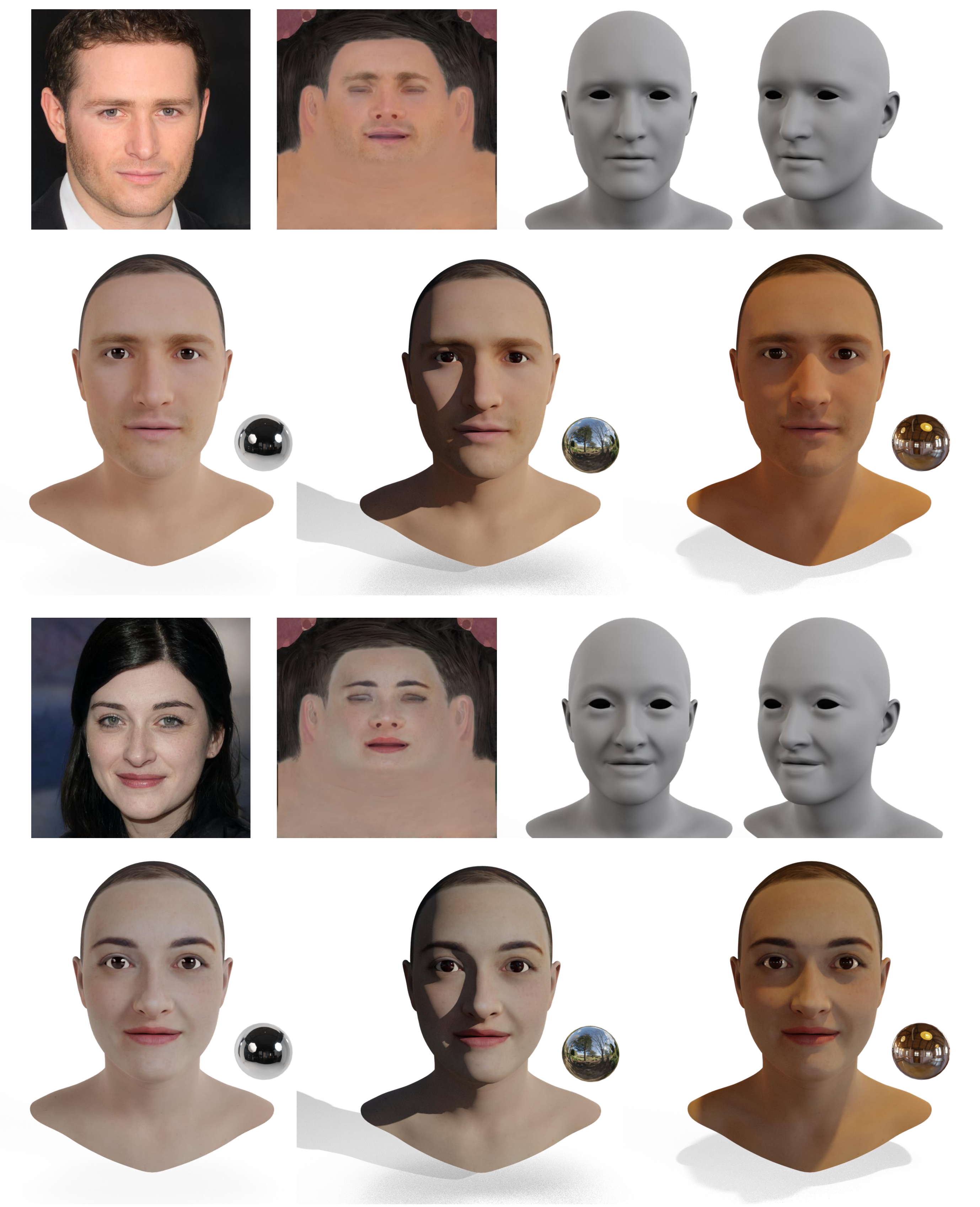}
   \caption{Examples of our reconstructed texture UV-maps, shapes, and renderings, where the produced textures are of high quality and without shadows which can be rendered with different lighting conditions.}
   \label{supp:fig:recons-relight-1}
\end{figure*}

\begin{figure*}[!t]
  \centering
   \includegraphics[width=0.95\linewidth]{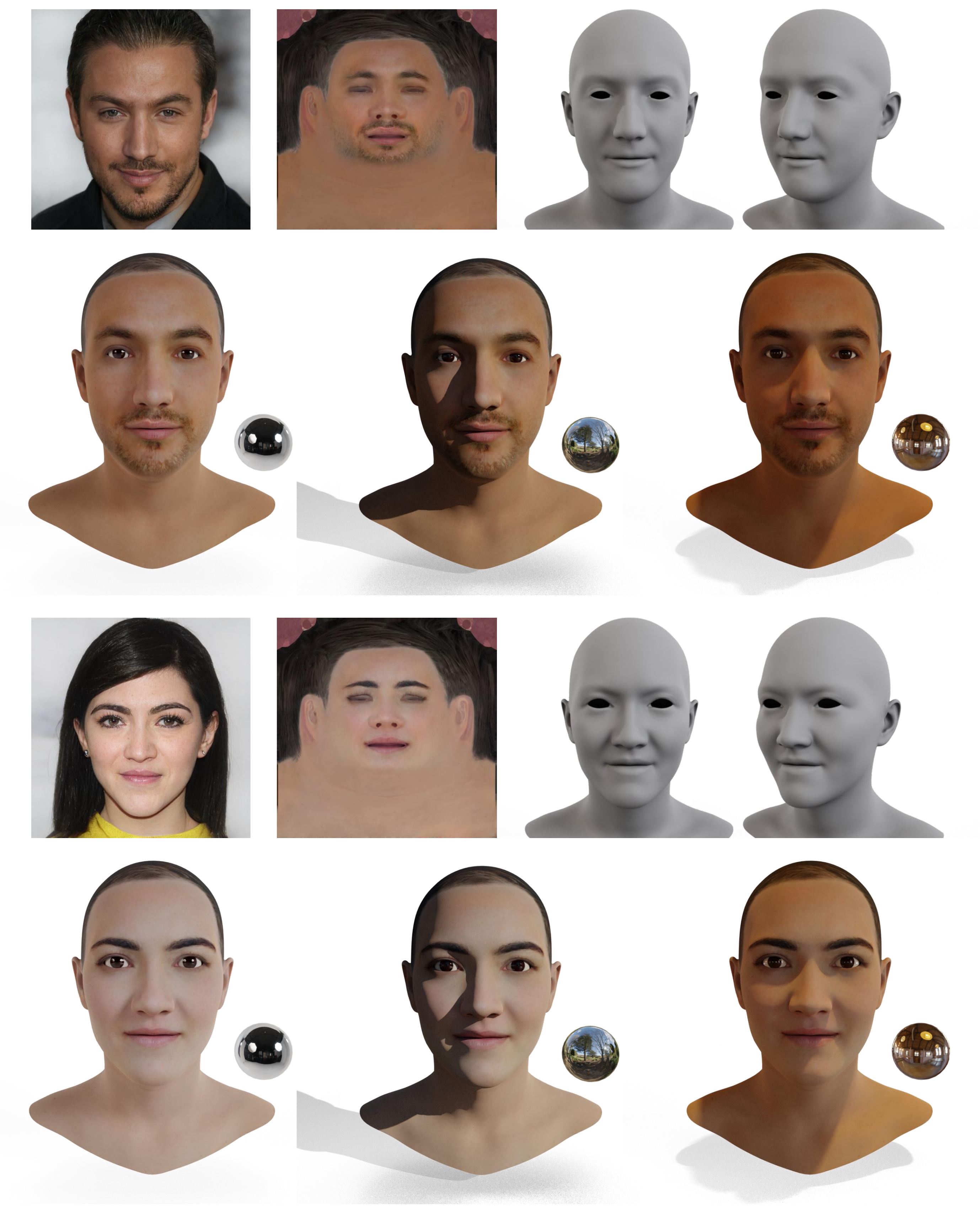}
   \caption{Examples of our reconstructed texture UV-maps, shapes, and renderings, where the produced textures are of high quality and without shadows which can be rendered with different lighting conditions.}
   \label{supp:fig:recons-relight-2}
\end{figure*}

\begin{figure*}[!t]
  \centering
   \includegraphics[width=0.95\linewidth]{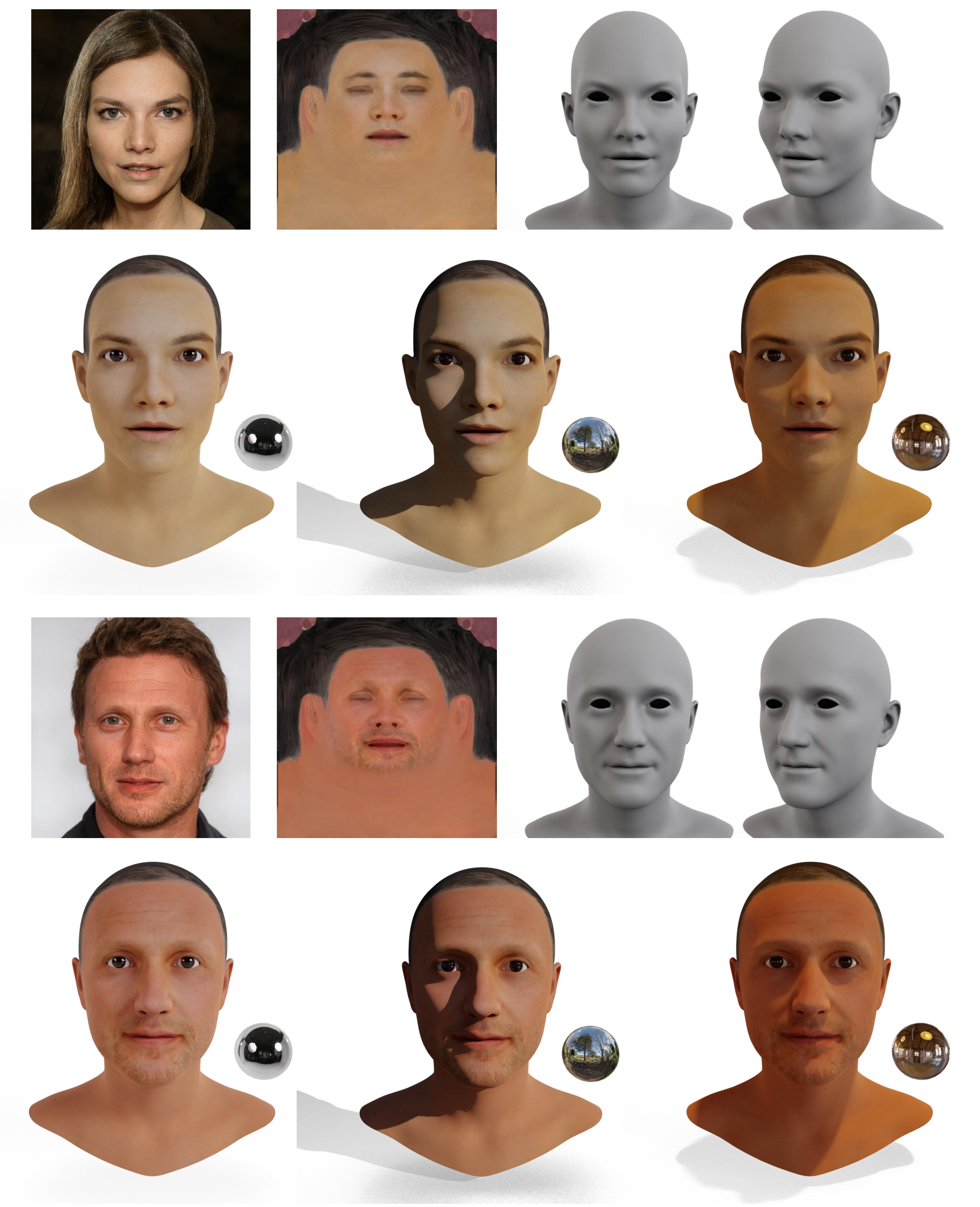}
   \caption{Examples of our reconstructed texture UV-maps, shapes, and renderings, where the produced textures are of high quality and without shadows which can be rendered with different lighting conditions.}
   \label{supp:fig:recons-relight-3}
\end{figure*}

In Fig.~\ref{supp:fig:recons-relight-1}-\ref{supp:fig:recons-relight-3}, we further show more examples of our reconstructed texture UV-maps, shapes, and renderings under different lighting conditions.

\section{More examples of FFHQ-UV dataset}
\label{supp:sec:unwrap-cases}

\begin{figure*}[!t]
  \centering
   \includegraphics[width=0.9\linewidth]{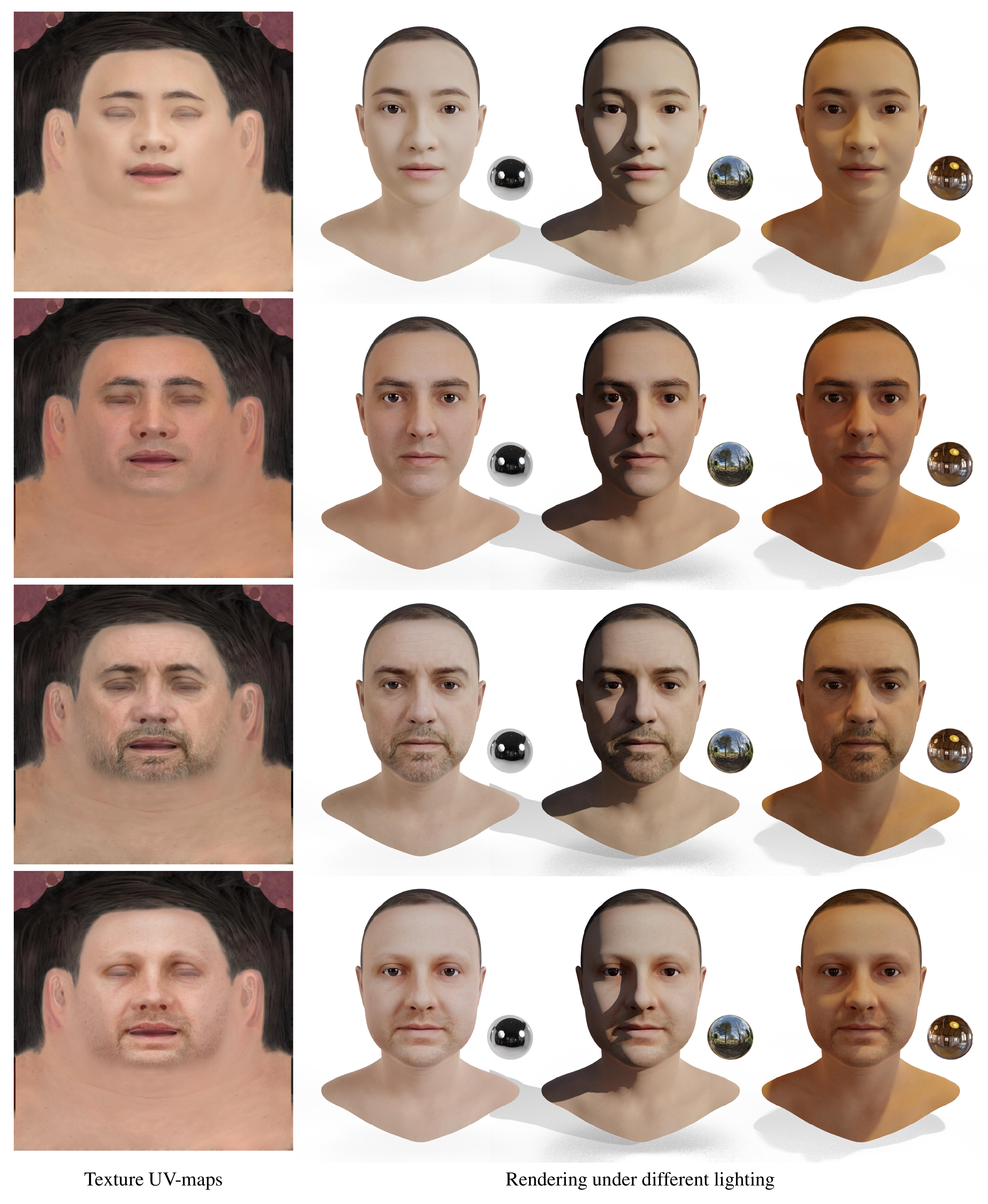}
   \caption{Examples of the proposed FFHQ-UV dataset, which are with even illuminations, neutral expressions, and cleaned facial regions (e.g., no eyeglasses and hair), and are ready for realistic renderings.}
   \label{supp:fig:unwrap-uv-1}
\end{figure*}

\begin{figure*}[!t]
  \centering
   \includegraphics[width=0.9\linewidth]{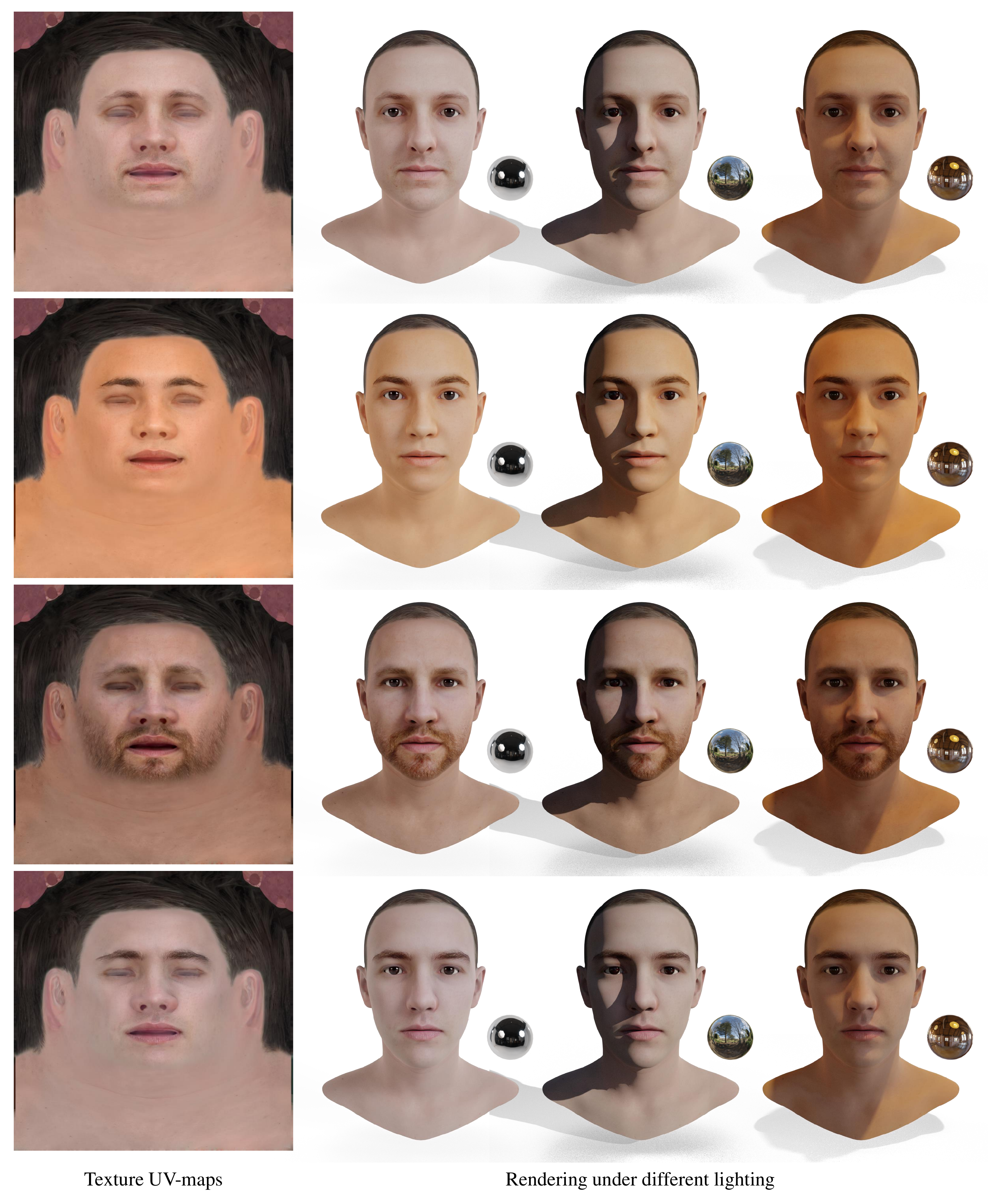}
   \caption{Examples of the proposed FFHQ-UV dataset, which are with even illuminations, neutral expressions, and cleaned facial regions (e.g., no eyeglasses and hair), and are ready for realistic renderings.}
   \label{supp:fig:unwrap-uv-2}
\end{figure*}

In this section, we provide more examples of the proposed FFHQ-UV dataset in Fig.~\ref{supp:fig:unwrap-uv-1}-\ref{supp:fig:unwrap-uv-2}.
The produced texture UV-maps are with even illuminations, neutral expressions, and cleaned facial regions (e.g., no eyeglasses and hair).
Thus, they are ready for rendering under different lighting conditions.
In our visualizations, there realistic environment lighting conditions are demonstrated including \href{https://polyhaven.com/a/studio_small_09}{a studio scene}\footnote{\url{https://polyhaven.com/a/studio_small_09}},
\href{https://polyhaven.com/a/garden_nook}{a garden scene}\footnote{\url{https://polyhaven.com/a/garden_nook}},
and \href{https://polyhaven.com/a/thatch_chapel}{a chapel scene}\footnote{\url{https://polyhaven.com/a/thatch_chapel}}).

{\small
\bibliographystyle{ieee_fullname}
\bibliography{egbib}
}